# Skillful Subseasonal-to-Seasonal Forecasting of Extreme Events with a Multi-Sphere Coupled Probabilistic Model

Bin Mu[1]†, Yuxuan Chen[1]†, Shijin Yuan[1]*, Bo Qin[1,2]*, Hao Guo[1]

[1]School of Computer Science and Technology, Tongji University, Shanghai, China.
[2]Department of Atmospheric and Oceanic Sciences & Institute of Atmospheric Sciences, Fudan University, Shanghai, China.

*Corresponding author. Email: yuanshijin@tongji.edu.cn (S.Y.); brunoqin@163.com (B.Q.)
†These authors contributed equally to this work.

## Abstract

Accurate subseasonal-to-seasonal (S2S) prediction of extreme events is critical for resource planning and disaster mitigation under accelerating climate change. However, such predictions remain challenging due to complex multi-sphere interactions and intrinsic atmospheric uncertainty. Here we present TianXing-S2S, a multi-sphere coupled probabilistic model for global S2S daily ensemble forecast. TianXing-S2S first encodes diverse multi-sphere predictors into a compact latent space, then employs a diffusion model to generate daily ensemble forecasts. A novel coupling module based on optimal transport (OT) is incorporated in the denoiser to optimize the interactions between atmospheric and multi-sphere boundary conditions. Across key atmospheric variables, TianXing-S2S outperforms both the European Centre for Medium-Range Weather Forecasts (ECMWF) S2S system and FuXi-S2S in 45-day daily-mean ensemble forecasts at 1.5°×1.5° resolution. Our model achieves skillful subseasonal prediction of extreme events including heat waves and anomalous precipitation, identifying soil moisture as a critical precursor signal. Furthermore, we demonstrate that TianXing-S2S can generate stable rollout forecasts up to 180 days, establishing a robust framework for S2S research in a warming world.

# Introduction

Deep neural networks have achieved remarkable progress in weather forecasting within 10 days by learning from reanalysis data, demonstrating superior performance compared to traditional numerical models across key statistical metrics (*1–6*). Our team has contributed to this advancement through TianXing (*6*), a model leveraging linear attention and physical decay mechanisms, which has achieved notable success in medium-range forecasting. The next frontier lies in extending these predictive capabilities to subseasonal-to-seasonal (S2S) timescales, typically spanning 2 weeks to 2 months, which holds critical importance for strategic resource allocation and early extreme hazard prevention (*7–9*).

Building upon the modeling approaches in weather forecasting, several pioneering studies have shown significant potential in artificial intelligence (AI) based S2S prediction. Current research primarily encompasses three technical strategies, including sophisticated model architectures that capture multiscale atmospheric dynamics (*10–13*), specialized training strategies that focus on long-term accuracy (*14*), and ensemble methods that quantify extend-range forecast uncertainties (*15–18*). A remarkable advance among them is FuXi-S2S (*16*), which utilizes a learnable perturbation block for S2S probabilistic modeling, representing the first data-driven method to outperform the ensemble mean forecast from the European Centre for Medium-Range Weather Forecasts (ECMWF) S2S system at 3- to 6-week lead times.

However, predicting extreme events in S2S timescale remains a substantial challenge. Specifically, deterministic S2S models inherently produce smoothed predictions due to optimization constraints, while existing probabilistic approaches primarily optimize ensemble mean statistics, failing to adequately capture extreme tail distributions that matter most for societal impacts (*19*). With climate change intensifying and extreme events becoming more frequent (*20*), novel methods are urgently needed to overcome this limitation. Therefore, we propose two complementary pathways for advancing S2S extreme event prediction:

1) Customizing multi-sphere coupling strategies. S2S forecasts face inherently low predictability due to their dependence on both rapidly decaying initial conditions and slowly varying boundary conditions (*21–25*). Incorporating multi-sphere boundary conditions is thus critical to constrain the atmospheric forecast state space and reduce climatology drift, thereby improving extreme event predictability (*7*). However, this requirement introduces a fundamental dilemma. On the one hand, predictors from different spheres exhibit fundamentally distinct internal variabilities. For instance, oceanic and land surface temperatures evolve orders of magnitude slower than atmospheric temperatures, suggesting the necessity to introduce sphere-specific feature extraction method rather than uniform approaches. On the other hand, the complex interaction processes between different spheres are only partly resolved, with many physical mechanisms remaining elusive (*26, 27*). From a physical perspective, these knowledge gaps underscore the need for physics-guided interaction modules that assist the model in learning physically meaningful multi-sphere interactions rather than data-driven artifacts. Despite recent efforts to incorporate multi-sphere variables, existing approaches are limited by either overly simplistic concatenation of all inputs, or purely end-to-end learning mechanisms that lack physical grounding (*28*). Fortunately, recent advances in feature engineering and optimization theory offer promising solutions. Deep neural network-based nonlinear autoencoders demonstrate powerful capability in extracting the distinct physical characteristics inherent to each sphere variable. Meanwhile, the development of optimal transport (OT) theory (*29*) provides a principled framework for treating boundary conditions and atmospheric variables as probability distributions, identifying mathematically optimal transfer pathways between them.

2) Developing well calibrated ensemble methods. According to Lorenz's theory of atmospheric predictability, forecast errors are unavoidable, and the sources of forecast uncertainty at S2S timescales are considerably more complex (*30*). Consequently, ensemble-based probabilistic forecasting is essential for accurately representing forecast distributions, particularly for capturing tail-end extreme events. Neural network-based forecasting has similarly embraced ensemble and uncertainty quantification techniques. A representative example is the variational autoencoder (VAE) framework employed by FuXi-S2S (*16*), which can capture large-scale uncertainty to some extent. However, limited by model bias, this approach of perturbing hidden features only once within the network leads to overly smooth forecasts and insufficient ensemble spread. Alternatively, diffusion models have emerged as a powerful tool for uncertainty modeling by solving stochastic differential equations (SDEs) through iterative denoising steps, which enables the generation of more diverse ensemble members (*31*, *32*). To our knowledge, few studies to date have explored diffusion models for uncertainty representation in S2S forecasts.

In this study, we combine these two aspects and introduce TianXing-S2S, a probabilistic model that provides global S2S daily-mean ensemble forecasts. TianXing-S2S integrates 81 variables from the atmosphere, ocean, land, and interface fluxes, substantially expanding both the number of predictors and the diversity of Earth system spheres compared to existing S2S models. Variables from each sphere are separately embedded into a compact latent space via a unique vector-quantized VAE (VQVAEs) (*33*), which enables sphere-specific feature extraction. In this latent space, a diffusion model is employed to model forecast uncertainty and generate daily forecasts, producing ensemble distributions that capture extreme events as comprehensively as possible. To achieve effective multi-sphere coupling, we propose the optimal transport block (OTB), a novel module within the denoiser network that facilitates physically meaningful information exchange between atmospheric variables and multi-sphere boundary conditions from a distribution transformation perspective. The incorporation of OTB significantly enhances the forecast skill, as demonstrated in ablation experiments in "Results". Through an autoregressive strategy, TianXing-S2S produces 45-day global daily-mean forecasts at $1.5°\times1.5°$ resolution with 51 ensemble members. Evaluations reveal that TianXing-S2S surpasses ECMWF S2S and FuXi-S2S across key atmospheric variables (e.g., geopotential, temperature, and humidity), with well-calibrated ensemble spread capturing subseasonal forecast uncertainty. Meanwhile, TianXing-S2S accurately captures the occurrence of extreme events, including mega heat waves and anomalous precipitation at 4-week lead times, identifying soil moisture feedback processes as key precursor signals for such successful prediction. Furthermore, we apply TianXing-S2S to long-term simulations by rolling out up to 180 days, the results maintain stable forecast modes and accurately reproduce seasonal cycles. These advances establish TianXing-S2S as a significant breakthrough toward reliable S2S prediction in a warming climate.

## Results

This study assesses the performance of TianXing-S2S using ERA5 reanalysis data spanning from 2018 to 2021 (*34*). To validate S2S prediction skill, TianXing-S2S autoregressively generates global 45-day daily mean forecasts with 51 ensemble members. The benchmark comprises two cutting-edge systems, including the ECMWF S2S reforecast (CY48r1) with 11 ensemble members at 46-day leads and the data-driven model FuXi-S2S with 51 members at 42-day leads. To ensure a fair comparison, all models are initialized on identical dates corresponding to the ECMWF S2S reforecast schedule and aligned to 1.5° resolution. The "Results" section is organized as follows. We first compare overall skills statistically through probabilistic metrics on ensemble distributions (in Fig. 1), deterministic metrics on ensemble means (in Fig. 1), and extreme metrics on tail-end coverage (in Fig. 2). Subsequently, six cases of anomalous temperature (4 cases) and precipitation (2 cases) events over East Asia are selected as examples to qualitatively evaluate the prediction capability for extreme events (in Fig. 3). Furthermore, attribution analysis is conducted to reveal the underlying physical processes governing TianXing-S2S's heat wave and precipitation prediction (in Fig. 4). Finally, a series of ablation experiments demonstrate the effectiveness of the proposed coupling block OTB (in Fig. 5).

Additional assessments including the skill on Madden Julian Oscillation (MJO) and long-term simulation (rollouts up to 180 days) stability are available in the Supplementary Figs. S1 to S3.

**Probabilistic and deterministic skills**

Ensemble forecasting is commonly assessed through statistical evaluations of probabilistic ensemble distributions and deterministic ensemble means (*35*). Fig. 1 compares TianXing-S2S with ECMWF S2S and FuXi-S2S for selected variables across different forecast lead times, averaged over the 2018 to 2021 test period. We employ the continuous ranked probability score (CRPS) and spread-skill ratio (SSR) to evaluate ensemble distributions, alongside weighted root mean square error (RMSE) and anomaly correlation coefficient (ACC) to assess ensemble mean accuracy.

TianXing-S2S demonstrates superior probabilistic predictive performance, consistently outperforming ECMWF S2S across most forecast variables with CRPS improvements of reaching up to 12% beyond 10 days (Fig. 1, A to F). The scorecard in Supplementary Fig. S4 reveals that TianXing-S2S generally surpasses the data-driven FuXi-S2S across most variables, with comparable performance observed only for specific variables such as geopotential, mean sea level pressure (MSLP), and total precipitation (TP) in lead times beyond 30 days. Furthermore, TianXing-S2S exhibits significantly superior ensemble reliability compared to FuXi-S2S, maintaining SSR values closer to the theoretical optimum of 1 (*36*), while also outperforms ECMWF S2S for several critical variables including geopotential at 500 hPa (Z500), 2-meter temperature (T2M), and TP (Fig. 1, G to J).

For deterministic metrics, ensemble mean evaluations (Fig. 1, K to N for RMSE and O to R for ACC) show that TianXing-S2S achieves notably lower RMSE compared to both benchmark models at 11 to 42 day lead times. For ACC, our model demonstrates leading performance for T2M beyond 2 weeks and superior skill for Z500, outgoing longwave radiation (OLR), and TP at 11 to 30 day leads, with performance comparable to FuXi-S2S at longer lead times. Comprehensive evaluations across additional meteorological variables, detailed in Supplementary Figs. S5 and S6, further validate TianXing-S2S's enhanced forecast skill and consistent performance advantages.

**Extreme skills**

Extreme events naturally manifest in the tail regions of statistical distributions, representing low-probability yet high-impact outcomes. Therefore, this subsection analyzes the predictive capability of TianXing-S2S for extreme values of two vital disaster-indicative variables T2M of heat wave and TP of precipitation.

Specifically, the 95th and 99th percentiles are selected as thresholds for Brier skill score (BSS) to determine extreme condition occurrence at each global grid point. The first row of Fig. 2 (Fig. 2, A and B) presents the weighted global mean BSS for ECMWF S2S, FuXi-S2S, and TianXing-S2S with different forecast lead times averaged across 2018 to 2021, where solid lines represent BSS values for the 95th percentile threshold and dashed lines for the 99th percentile. Beyond 10 days after forecast initialization, TianXing-S2S demonstrates significant advantages in extreme values prediction, achieving optimal BSS scores for both T2M and TP compared to the benchmarks. Notably, for extreme high-temperature events (dashed lines in Fig. 2 B), TianXing-S2S maintains positive BSS scores, creating a considerable performance gap compared to ECMWF S2S and FuXi-S2S, whose predictive skill progressively degrades below zero after 10-day lead times.

The subsequent two rows of Fig. 2 further illustrate the global distribution of BSS for T2M (Fig. 2, C to E) and TP (Fig. 2, F to H). These distributions are obtained by averaging daily global BSS values across weeks 3 to 6. The three columns from left to right show ECMWF S2S, TianXing-S2S, and their difference (with red indicating superior performance by TianXing-S2S and blue indicating the opposite). Regarding T2M, TianXing-S2S achieves enhanced scores across the Northeast Pacific, South Pacific, and tropical Indian Ocean basins, while showing comparative deficiencies in the North Pacific and some regions of the Southern Hemisphere oceans. Regarding precipitation forecasts, TianXing-S2S exhibits substantially enhanced skill relative to the numerical model across virtually all regions, with only limited areas in North Africa showing comparable or inferior performance. Additionally, TianXing-S2S shows exceptional performance over continental areas, with the difference maps in the third column displaying almost exclusively positive values over land. This suggests that the inclusion of land surface variables may facilitate better representation of land-atmosphere interactions (*37–39*), ultimately leading to more skillful near-surface forecasts.

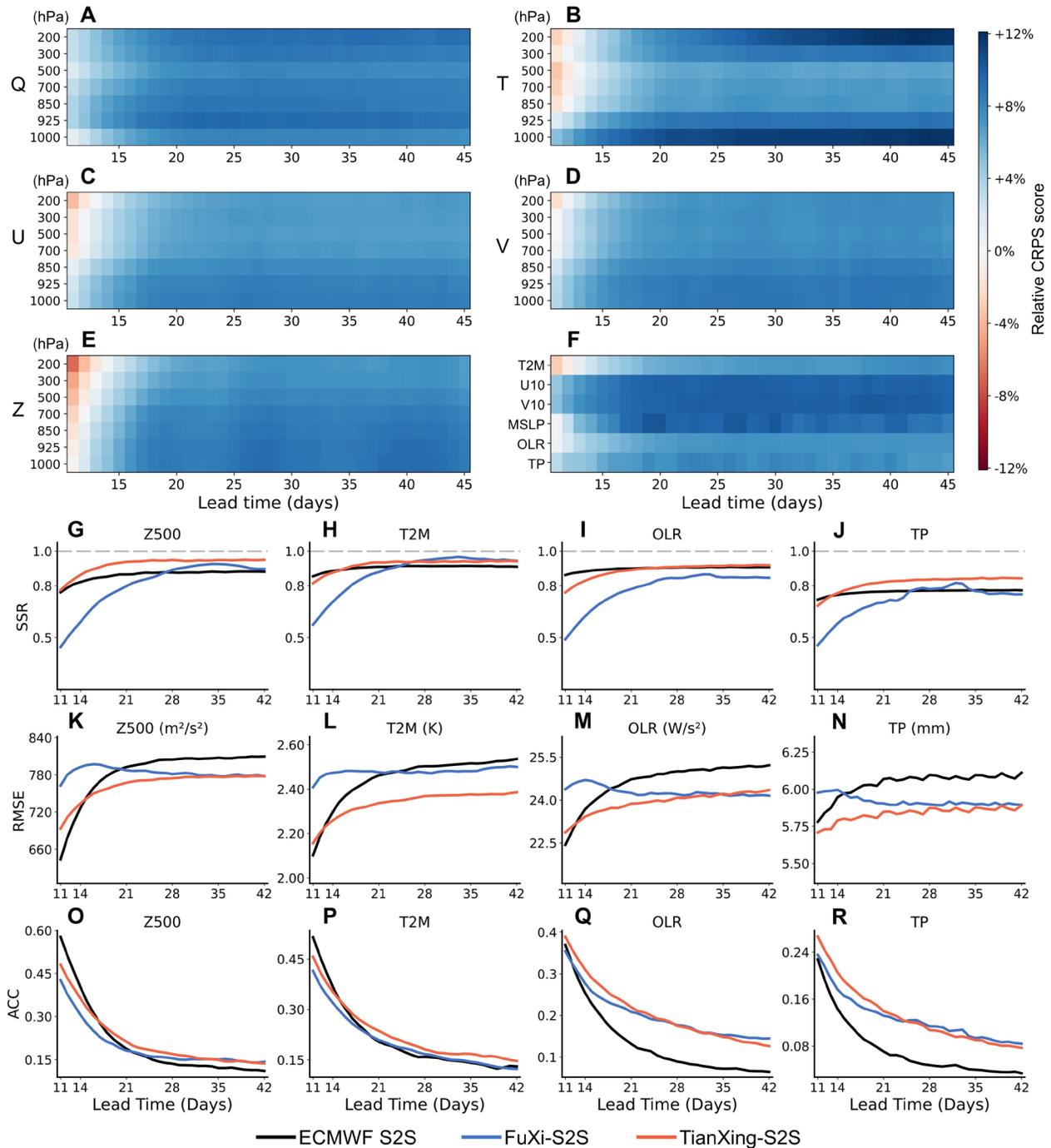

**Fig. 1. Probabilistic and deterministic evaluations of TianXing-S2S against benchmark models ECMWF S2S and FuXi-S2S.** (**A** to **F**) CRPS scorecards comparing TianXing-S2S ensemble versus ECMWF S2S ensemble. Shown are relative CRPS score as function of lead time (day 11 to 42) for all atmospheric variables. Blue colors mark score improvements and red colors mark score degradations of TianXing-S2S. (**G** to **J**) SSR comparison across lead times of ECMWF S2S, FuXi-S2S, and TianXing-S2S for geopotential at 500 hPa (Z500), 2-meter temperature (T2M), outgoing longwave radiation (OLR) and total precipitation (TP). The dashed line at 1 denotes the theoretical optimum. (**K** to **N**) Same as (G to J), but for RMSE comparison of ensemble mean. (**O** to **R**) Same as (G to J), but for ACC comparison of ensemble mean. All metrics are derived from test samples averaged over 2018 to 2021.

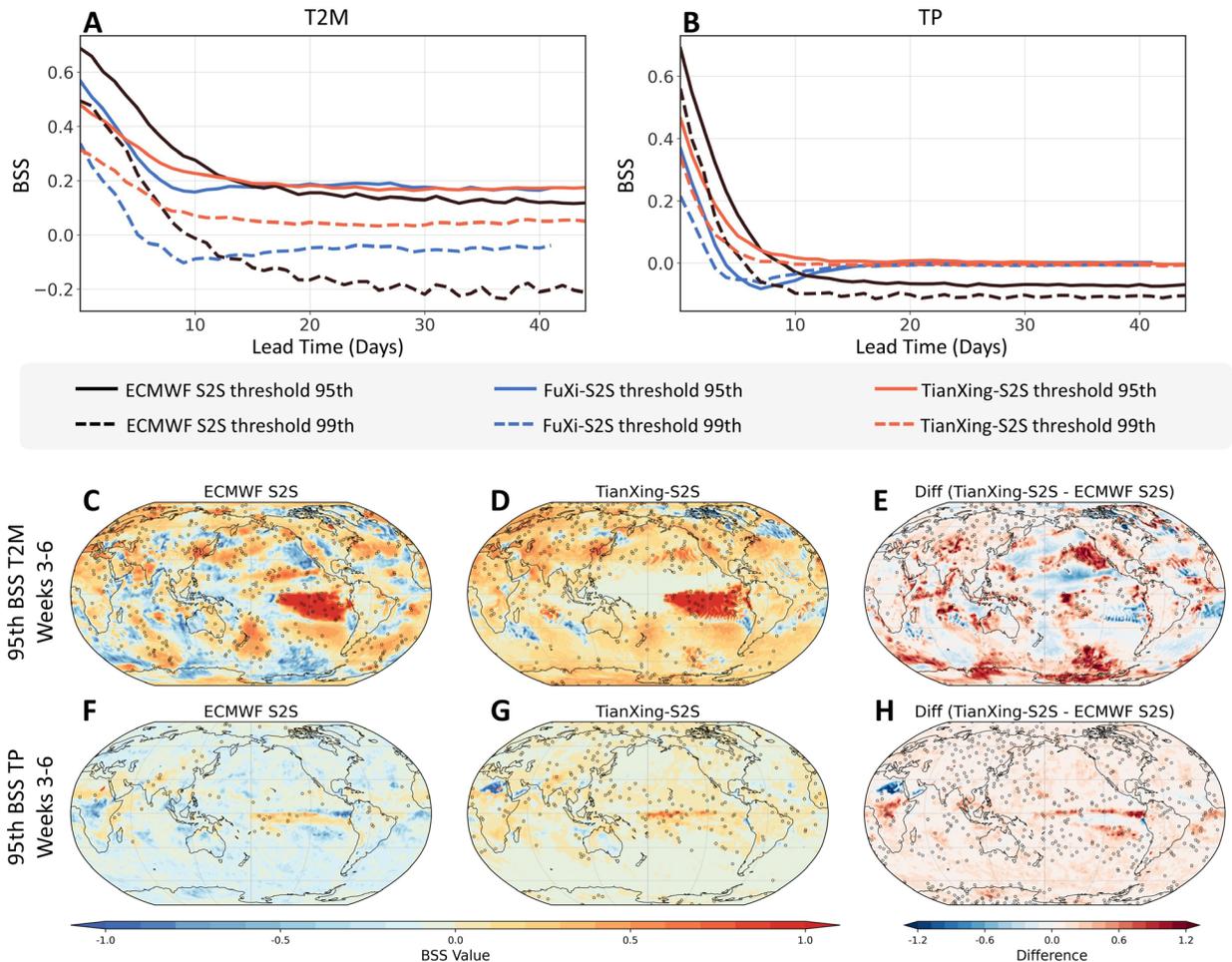

**Fig. 2. Extreme skill of TianXing-S2S assessed by BSS for T2M and TP, benchmarking against ECMWF S2S and FuXi-S2S.** (**A** and **B**) Global mean BSS at 95th and 99th percentile thresholds for T2M (A) and TP (B) across lead times (day 1 to 45). (**C** to **H**) Spatial patterns of BSS at the 95th percentile threshold, showing week 3 to 6 averages for T2M (C to E) and TP (F to H). Columns from the left to right show spatial BSS for ECMWF S2S, TianXing-S2S, and their difference, respectively. Blue colors mark score improvements and red colors mark score degradations of TianXing-S2S. Gray circles mark grid points exceeding the 95% confidence level (two-tailed t-test), spatially subsampled for clarity. The BSS values are derived from test samples averaged over 2018 to 2021.

### East Asian extreme heat wave and precipitation predictions

As one of the world's most climatically dynamic regions, East Asia has experienced increasing frequency of extreme weather events under ongoing climate change (*40*). We conduct six extreme

6 | 38

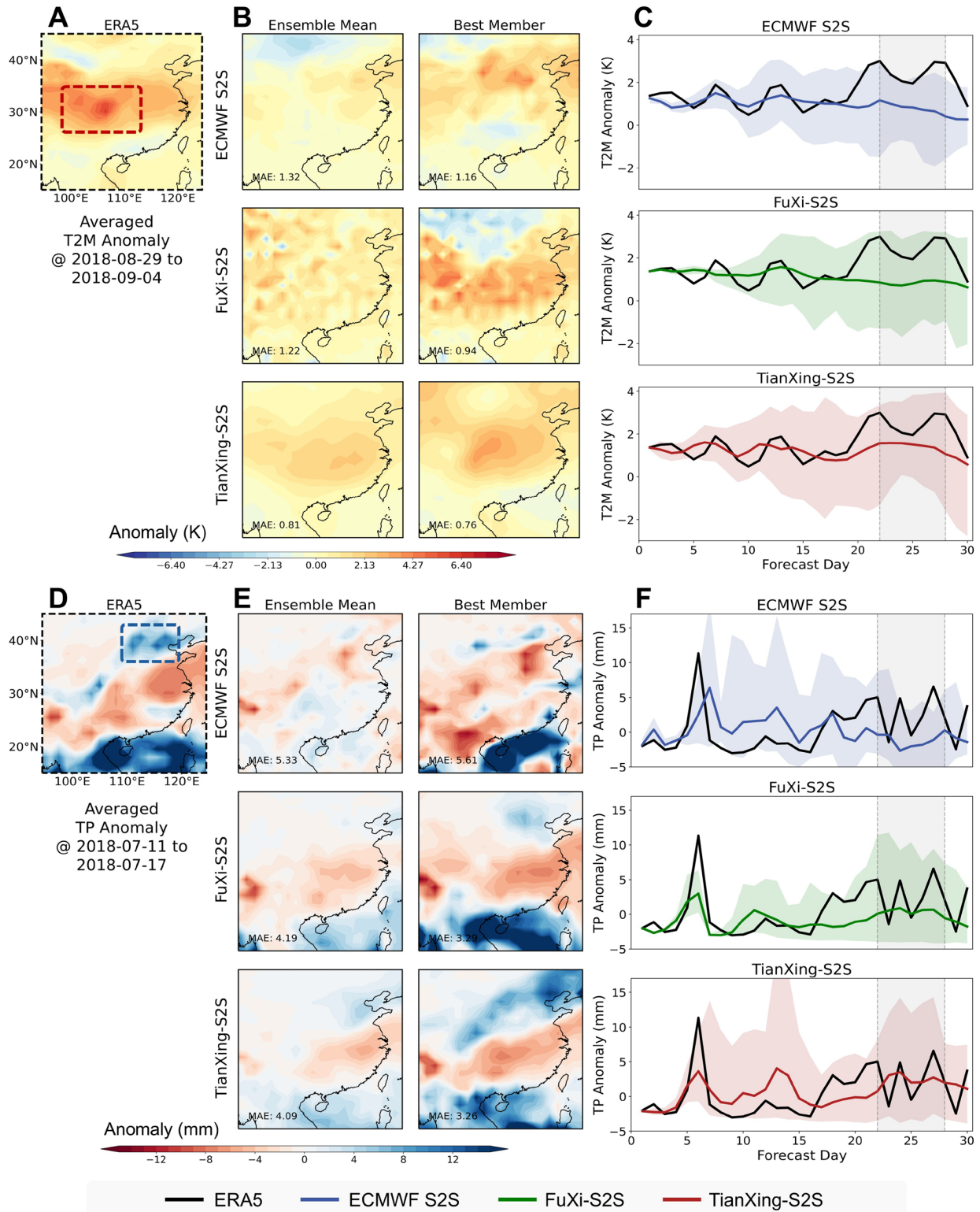

**Fig. 3. Comparison of 4-week lead predictions for East Asian extreme heat wave and precipitation events among TianXing-S2S, FuXi-S2S, and ECMWF S2S. (A** to **C)** Heat wave event over the Sichuan Basin during August 29 to September 4, 2018. (A) ERA5 averaged T2M anomalies as ground truth, with red dashed box showing heat wave region. (B) Ensemble mean (left column) and best-performing member (right column) predictions at 4-week leads of ECMWF S2S (top row), FuXi-S2S (middle row), and TianXing-S2S (bottom row). Numbers in lower left corners indicate domain-averaged absolute errors of each prediction. (C) Temporal evolution of area-averaged T2M anomalies over the boxed region in (A) for ECMWF S2S (top), FuXi-S2S (middle), and TianXing-S2S (bottom). Black lines denote ERA5 reanalysis, colored lines show ensemble means, and shading indicates ensemble spread. Gray boxes mark the extreme event breakthrough window. **(D** to **F)** Same as (A to C) but for the precipitation event over North China during July 11 to 17, 2018.

7 | 38

case studies over East Asia to evaluate the potential of TianXing-S2S for disaster early warning at 4-week lead times. Fig. 4 displays two examples for an extreme heat wave event (Fig. 4, A to C) and an extreme precipitation event (Fig. 4, D to F). Additional four simulations including heat waves, cold waves, and precipitation events are provided in Supplementary Figs. S7 to S10.

During August 29 to September 4, 2018, the Sichuan Basin experienced a severe heat wave event that brought record-breaking temperatures to this densely populated region (Fig. 4 A). We employ TianXing-S2S to reforecast (also termed hindcast) this temperature extreme by a 51-member ensemble initialized 4 weeks earlier, benchmarking against the 11-member ECMWF S2S reforecast and the 51-member FuXi-S2S hindcast. Among all models, TianXing-S2S demonstrates exceptional capability in capturing the heat wave's spatial patterns through both ensemble mean and optimal ensemble members, particularly exhibiting high fidelity for temperature anomaly distributions over the Sichuan basin (Fig. 4 B). ECMWF S2S fails to predict meaningful temperature anomalies, while FuXi-S2S only partially captures positive temperature signals with degraded spatial coherence and displaced maximum heating centers. In terms of quantitative accuracy, TianXing-S2S achieves the smallest absolute error, confirming its effectiveness in simulating this extreme heat event.

Furthermore, we quantify the evolution of ensemble distributions at different lead times across the heat wave's epicenter (Fig. 4 C). The black solid line represents ERA5 reanalysis, colored solid lines denote ensemble means, and shaded regions indicate the spread of ensemble members. Gray boxes delineate the temporal window of extreme event occurrence, corresponding to forecast week 4. Comparison of ensemble distributions reveals distinct characteristics across models. ECMWF S2S exhibits reasonable spread during the first three weeks but shows limited dispersion in week 4, failing to capture the extreme temperature magnitudes of the heat wave. We attribute this limitation partly to its constrained ensemble size of 11 members. FuXi-S2S, employing a larger ensemble of 51 members, successfully encompasses the extreme temperatures. However, it demonstrates notably narrow spread during the initial forecast period (approximately days 1 to 15), suggesting that the VAE-based perturbation approach relies predominantly on error accumulation through iterative forecasting rather than learning to generate physically consistent ensemble spread from initialization. In contrast, TianXing-S2S produces the most skillful ensemble forecasts in both mean and spread, generating members that achieve both broad coverage and high accuracy throughout nearly the entire forecast window. These results demonstrate that diffusion models substantially outperform VAE-based methods in representing S2S forecast uncertainty, establishing their effectiveness as an ensemble generation framework. Visualization of randomly selected ensemble members in Supplementary Fig. S11 reveals the diversity and high dispersion of TianXing-S2S ensemble patterns, further validating our model's advantages in ensemble forecasting of extreme events.

The second case examines extreme summer precipitation over North China during July 11 to 17, 2018. The ERA5 reanalysis reveals a meridional tripole pattern (Fig. 4 D), where North China experienced enhanced rainfall from the summer monsoon rain belt, the Yangtze River basin underwent post-Meiyu drought conditions, and South China witnessed its secondary rain belt, collectively forming a distinctive "+−+" anomaly distribution. At 4 weeks lead time (Fig. 4 E), three models can partially capture the central precipitation deficit in the Yangtze River basin, yet only TianXing-S2S correctly predicts the complete tripole structure in its ensemble mean, including the crucial northern rain belt. Similarly, the ensemble spread coverage (Fig. 4 F) and visualization of ensemble members (Supplementary Fig. S12) consistently validate TianXing-S2S's exceptional capability relative to both traditional numerical models and current state-of-the-art data-driven approaches FuXi-S2S.

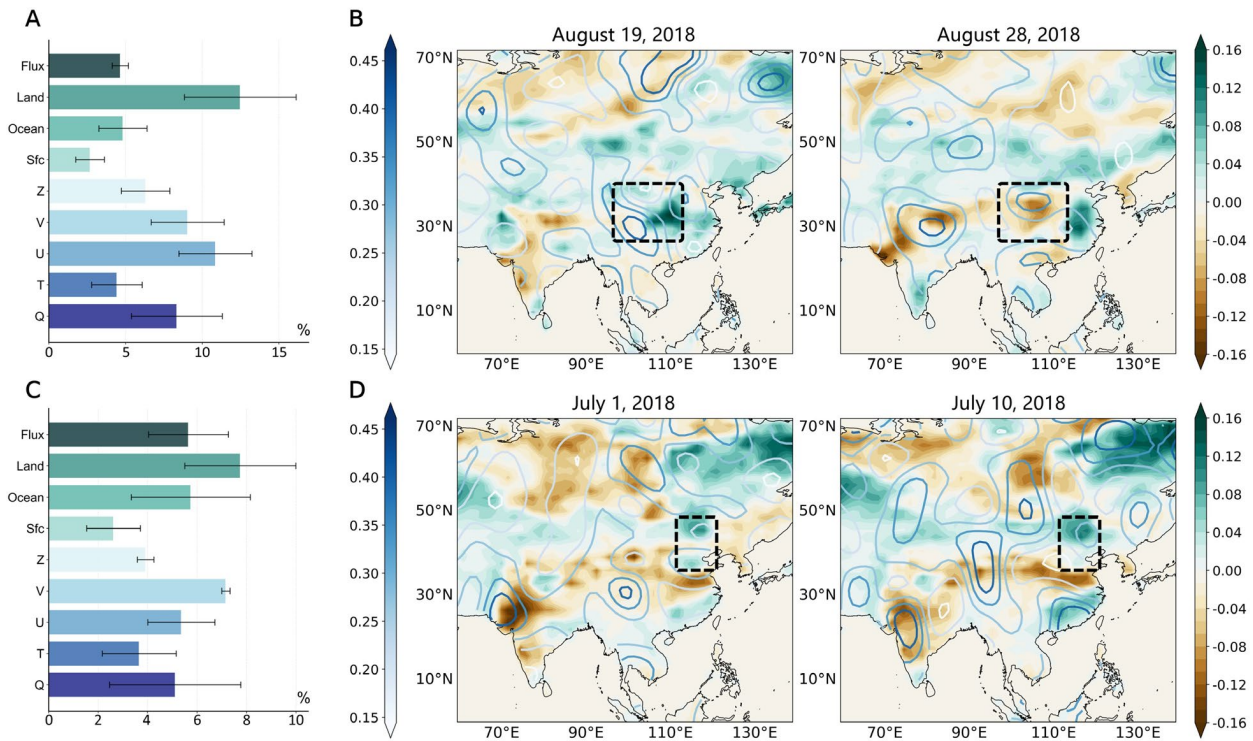

**Fig. 4. Multi-sphere predictor attribution and soil moisture precursor signals identified by TianXing-S2S for East Asian extreme events prediction.** First and second rows show the results for heat wave (August 29 to September 4, 2018) and precipitation event (July 11 to 17, 2018), respectively. **(A)** Attribution of multi-sphere predictors using the perturbation importance method (PIM). Error bars represent standard deviation across 51 ensemble members. **(B)** Saliency maps overlaid with soil moisture anomalies for the best-performing ensemble member at 10 days and 1 day before extreme event onset. Color shading indicates soil moisture anomalies (right colorbar) and contours show saliency values from GradCAM (left colorbar). Bluer contours indicate higher attention from TianXing-S2S. Dashed boxes delineate the extreme event region. **(C and D)** Same as (A and B) but for the precipitation event.

**Soil moisture precursor signals for extreme heat wave and precipitation events**

To investigate the physical basis underlying TianXing-S2S predictions and explain its capability to forecast extreme events, we conduct attribution analysis on the model's predictions for the East Asian extreme heat wave and precipitation events described in the previous subsection. The analysis proceeds in two steps. We first employ the perturbation importance method (PIM) (*41*) by shuffling initial input predictors and measuring the resulting increase in forecast errors, thereby quantifying the relative contribution of variables in different spheres to forecast skill. Subsequently, targeting the most influential variables identified by PIM, we utilize the gradient-weighted class activation mapping (GradCAM) method (*42*) to generate saliency maps that visualize the spatial patterns of TianXing-S2S's focus for these key variables.

Fig. 5 presents attribution analysis results of TianXing-S2S's predictions for the heat wave event (during August 29 to September 4, 2018, in subgraphs A and B) and the precipitation event (July 11 to 17, 2018, in subgraphs C and D). For each event, we partition all input variables into six atmospheric variable groups and three boundary condition groups. The atmospheric groups comprise five upper-air variables at 13 pressure levels and surface variables (denoted as Sfc). The three boundary condition groups comprise corresponding ocean, land, and flux variables, respectively. For each group, we randomly shuffle its values prior to model input, effectively disrupting the historical information from that group. The contribution of each group to model predictions is then assessed by measuring the resulting increase in forecast RMSE. For the heat wave event (Fig. 5 A), variables from land sphere exhibit the strongest influence among initial input fields, followed by wind components (U and V) and specific humidity (Q). Temperature variables show surprisingly modest influence despite their direct physical connection to heat extremes. This reveals that skillful subseasonal forecasting of extreme heat events requires capturing synergistic land-atmosphere interactions rather than relying solely on atmospheric temperature fields.

After identifying the importance of soil variables to forecast skill, we use the GradCAM method to compute saliency maps of week-4 T2M forecasts regard surface soil moisture of the best-performing ensemble member. Accounting for the slow-varying nature of soil moisture, we select these two snapshots at 10 days (August 19) and 1 day (August 28) prior to the heat wave onset to concisely illustrate the temporal evolution. Fig. 5 B displays the saliency maps overlaid with soil moisture anomalies at these two steps, bluer contours with higher saliency values indicate greater model focus. The black dashed boxes delineate the heat wave region. At 10-day lead time, incipient negative soil moisture anomalies appeared over the Sichuan Basin. By one day before the heat wave event, the Sichuan Basin exhibits pronounced negative anomalies consistent with the heat wave pattern. TianXing-S2S identified these dry anomalies as key precursor signals throughout the heat wave prediction, with contours covering the dryness anomaly regions. We infer that this progression reflects land-atmosphere feedback mechanisms during heat wave development (*43*). Specifically, depleted soil moisture reduces evapotranspiration and its associated cooling effect, leading to elevated surface temperatures. The resulting drier atmospheric conditions decrease cloud cover and enhance incoming solar radiation, creating a positive feedback loop that amplifies heat extremes. The discovery of precursor signals from soil moisture demonstrates that TianXing-S2S successfully captures physics-meaningful information during its inference.

Similarly, for the precipitation event over North China during July 11 to 17, 2018, land surface predictors again exhibit the strongest influence (Fig. 5 C), while atmospheric dynamical variables including winds and oceanic variables show moderate contributions. The saliency analysis in Fig. 5 D reveals that soil moisture anomalies from south to north displayed a tripole pattern consistent with the precipitation distribution at both time steps, while contours over North China closely match the anomaly pattern. We associate such precursor signals captured by TianXing-S2S to the positive feedback effect between soil moisture and precipitation, whereby soil moisture anomalies regulate atmospheric circulation and moisture convergence patterns, thereby enhancing forecast accuracy (*44*). These findings collectively indicate that TianXing-S2S identifies the critical role of soil moisture in extreme event prediction.

Supplementary Fig. S13 exhibits TianXing-S2S's accurate representation of multi-sphere boundary conditions through comparison with ERA5 reanalysis, a crucial verification as reliable simulation of these boundary conditions underpins the credibility of our attribution analysis above. Nevertheless, we acknowledge that the actual physical mechanisms are considerably more complex, particularly regarding the synergistic coupling and interactions among variables across different spheres, which merit more detailed investigation in future research.

**Ablation study of the OTB**

We conduct ablation experiments to verify the impact of the OTB on forecast accuracy. Specifically, we design four comparative experiments consisting of the ECMWF S2S ensemble forecast, a variant replacing denoiser with a simple convolutional network SimVP (*45*), a variant replacing the OTB with a cross-attention mechanism, and the full TianXing-S2S model. We use week-4 averaged CRPS scores for T2M and MSLP as evaluation metrics, with all models employing identical predictors and training procedures. Results in Fig. 5 A and B reveal that compared to the ECMWF S2S baseline, simple model SimVP struggles to surpass numerical model. The cross-attention-based variant demonstrates reasonable performance but remains inferior to the model incorporating the OTB, highlighting the substantial contribution of our proposed module to forecast accuracy improvement.

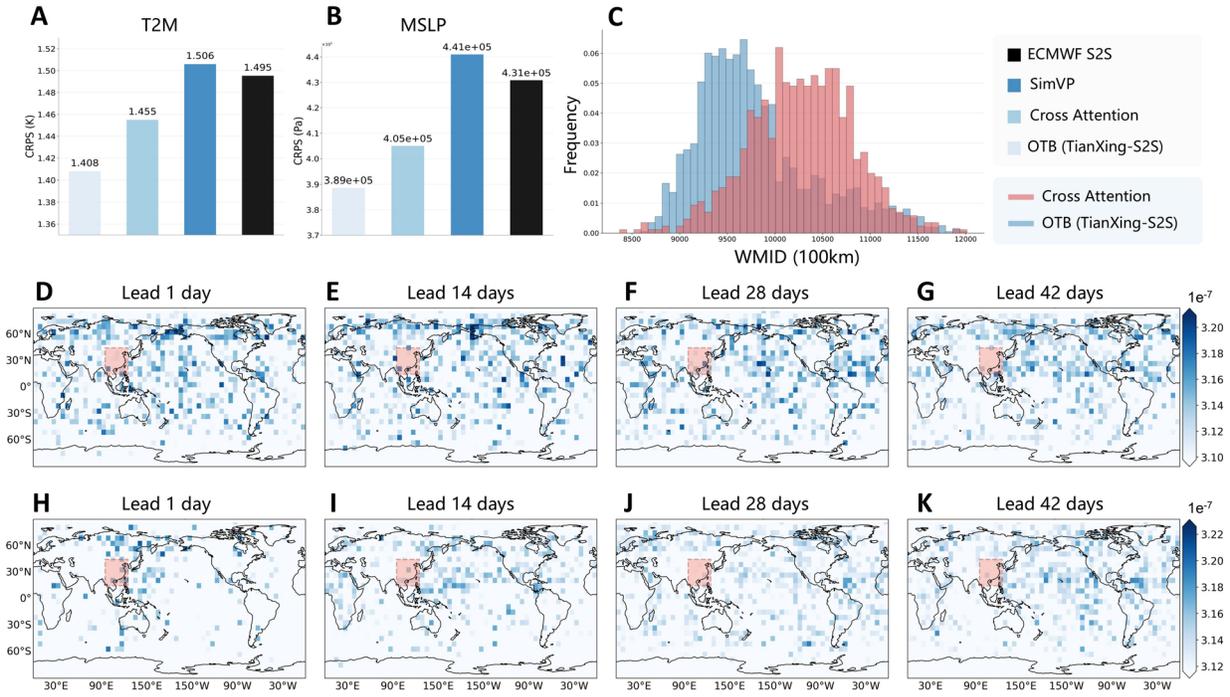

**Fig. 5. Ablation experiments of the OTB. (A** and **B)** Ablation results of four approaches, including ECMWF S2S, a variant replacing denoiser with a simple convolutional network SimVP, a variant replacing the OTB with cross-attention, and the full TianXing-S2S model. Performance is evaluated using week-4 averaged CRPS scores for T2M (A) and MSLP (B), averaged over 2018 to 2021. **(C)** Frequency distribution of weighted mean influencing distance (WMID) between global boundary conditions and East Asia of the OTB (blue) and cross-attention (red) blocks. Lower and broader WMID values indicate more physically consistent multi-sphere coupling according to predictability theory. **(D to K)** Spatial influences of global multi-sphere boundary conditions on East Asian identified by cross-attention (D to G) and OTB (H to K) at 1, 14, 28, and 42-day leads, averaged across test set. Red boxes delineate the East Asian region while blue shading indicates influence strength, with bluer indicating stronger influence. Influence values at each grid point are centered by subtracting the global mean and filtered to display only the top 50th percentile.

To further investigate the inference mechanism of the OTB, we examine the interaction between atmospheric predictands and global boundary conditions by visualizing the optimal transport matrix (OTM). Analogous to cross-attention matrices employed in (*13*), each element in the OTM quantifies the coupling strength between grid points in multi-sphere boundary conditions and atmospheric variables. Through appropriate matrix slicing and reshaping, we can trace the influence of global boundary conditions on atmospheric forecasts over specific regions, such as East Asia. Specifically, we archive the OTMs computed within the OTBs during inference on the test set and average them according to forecast lead time. For visual clarity, we present results at representative lead times of 1, 14, 28, and 42 days (Fig. 5, D to K). Each visualization matrix has been centered by subtracting its global mean, and we display only grid points with values in the top 50th percentile (Fig. 5, H to G). For comparison, we present the corresponding attention matrices from the cross-attention-based variant (Fig. 5, D to G). In each panel, red boxes delineate the East Asian region, while blue shading indicates the contribution of global boundary conditions to East Asian forecasts, with darker blue denoting stronger influence. Results reveal that the OTM initially focuses on proximate regions such as Siberia and the Northwest Pacific. As lead time increases, the attention progressively extends to encompass global scales, indicating a transition from local dependencies to remote predictability sources as forecasts enter the subseasonal range. This evolutionary behavior of the OTB aligns with theoretical understanding of subseasonal predictability sources (*46*). In contrast, the cross-attention matrices exhibit no systematic distributional evolution. Even at 1-day lead time, East Asian predictions depend extensively on Southern Hemisphere oceanic regions, potentially leading the model to exploit spurious correlations rather than learning physically consistent processes.

Moreover, we introduce the weighted mean influencing distance (WMID) metric, which quantifies the average distance between influential global boundary conditions and East Asia. Intuitively, WMID calculates the weighted average distance from blue grid points to the East Asian region presented in Fig. 5 D to K, providing quantitative verification of the observed expansion of predictability sources. The WMID frequency distribution across all test samples (Fig. 5 C) reveals that the OTB-based method exhibits overall smaller WMID values with a broader, right-skewed distribution. This suggests that predictability information captured by the OTB concentrates primarily near East Asia, progressively diminishes with distance, and ultimately spans the entire globe. In contrast, the cross-attention method displays a frequency distribution resembling a normal distribution, indicating a nearly uniform global dispersion of attention, which is statistically inconsistent with subseasonal predictability theory (23, 24). These results demonstrate that the OTB more effectively learns physically consistent coupling processes between atmospheric variables and multi-sphere boundary conditions.

## Discussion

In this study, we developed TianXing-S2S, a probabilistic model that advances data-driven global S2S daily-mean ensemble forecasting. The model integrates 81 variables spanning the atmosphere, ocean, land, and surface fluxes. This variable set is substantially larger than those used in current S2S models, enabling more comprehensive representation of multi-sphere interactions. TianXing-S2S first employs sphere-specific VQVAEs to encode diverse multi-sphere predictors into a compact latent space, where a diffusion framework then generates 51-member daily ensemble forecasts extending to 45 days. A key innovation is the OTB embedded within the denoiser, which optimizes interactions between atmospheric variables and multi-sphere boundary conditions through distribution transformation. This approach differs fundamentally from simple variable concatenation or purely end-to-end learning adopted by existing models. Comprehensive evaluations demonstrate that TianXing-S2S outperforms both the ECMWF S2S system and the state-of-the-art FuXi-S2S model. Improvements are observed across key meteorological variables in ensemble mean accuracy, ensemble spread, and tail-end distribution coverage. Furthermore, TianXing-S2S achieves superior skill in extreme event prediction, including heat waves, cold spells (Supplementary Figs. S8 and S9), and precipitation extremes at 4-week lead times. For the MJO, a prominent subseasonal phenomenon, TianXing-S2S demonstrates remarkable skill in reproducing its intrinsic convective and dynamical signatures (Supplementary Fig. S1). These prediction capabilities hold significant implications for enhancing societal resilience to weather extremes.

Moreover, TianXing-S2S demonstrates meaningful progress in physical interpretability. For predicting extreme heat waves and precipitation events over East Asia, TianXing-S2S effectively identifies soil moisture as a key precursor signal. Furthermore, through the OTB, TianXing-S2S exhibits physically consistent inference processes aligned with subseasonal predictability theory, with attention progressively expanding from local to global scales as forecast lead time increases, reflecting the transition from initial-condition-dependent to boundary-forced predictability at subseasonal timescales. These results suggest that integrating data-driven computational efficiency with physical priors represents a promising pathway toward more interpretable and skillful forecasting systems.

In addition to interpretability, extending predictive capability to longer timescales remains a critical frontier for data-driven models. Advances in S2S prediction serve as an essential bridge toward developing seamless weather-climate prediction systems (23, 24). The primary obstacle for AI models in this endeavor is preserving forecast stability, as they frequently suffer from mode collapse and severe climatological drift, typically failing to maintain viable forecasts beyond one season (*49*). TianXing-S2S demonstrates encouraging progress in addressing this stability challenge. In the long-term forecast experiments shown in Supplementary Figs. S2 and S3, TianXing-S2S runs autoregressively for forecasts up to 180 days, maintaining clear and stable modal structures for both atmospheric and multi-sphere boundary condition variables while accurately capturing seasonal cycles. This robust long-range stability establishes considerable potential for developing seasonal or even interannual seamless prediction systems based on AI frameworks.

Looking forward, numerous opportunities for improvement and exploration exist. Although TianXing-S2S incorporates variables from different spheres, the coupling remains incomplete, as each sphere operates on distinct timescales with unique internal variability rather than unified dynamics. Future research should develop specialized models for individual Earth system components and integrate them through dedicated couplers, thereby advancing toward data-driven Earth system models. Additionally, based on the two types of predictability theory (*50*), our generative model approach to ensemble forecasting represents the second type through parameter perturbations. We are optimistic that simultaneously incorporating both initial condition perturbations (first type) and parameter perturbations (second type) in data-driven models will further enhance ensemble forecast skill for S2S prediction.

## Materials and Methods

### Datasets

ERA5 represents the fifth-generation atmospheric reanalysis from the ECMWF, featuring approximately 31 km horizontal resolution and hourly temporal coverage from 1940 to present (*34*). Through sophisticated quality control and observational data assimilation techniques, ERA5 has been widely recognized as the highest-quality reanalysis product currently available. TianXing-S2S employs ERA5 as its exclusive dataset for model development.

For variable selection, accurate subseasonal prediction requires adequate representation of multi-sphere interactions, including tropical convection, stratosphere-troposphere coupling, air-sea exchanges, land-atmosphere feedbacks, and sea ice variability (*23*). Atmospheric variables alone are insufficient to capture the evolution of atmospheric dynamics at subseasonal timescales. To this end, this study incorporates multi-sphere variables from atmosphere, ocean, land surface, and interface fluxes as both input predictors and predictands for TianXing-S2S, totaling 81 variables. Specifically, within the atmospheric sphere, TianXing-S2S utilizes 71 variables comprising specific humidity (Q), temperature (T), u component of wind (U), v component of wind (V), and geopotential (Z) at 13 pressure levels (50, 100, 150, 200, 250, 300, 400, 500, 600, 700, 850, 925, and 1000 hPa), along with surface variables including 2-meter temperature (T2M), outgoing longwave radiation (OLR), total precipitation (TP), mean sea-level pressure (MSLP), 10-meter u wind component (U10), and 10-meter v wind component (V10). In the oceanic sphere, sea surface temperature (SST) and sea-ice cover (SIC) are included to represent ocean thermal state and cryosphere processes (*51*). Land surface soil temperature (ST) and moisture (SM) possess significant memory effects that modulate land-atmosphere heat and moisture transports (*52*), thus TianXing-S2S incorporates soil temperature and moisture at three depth layers (layer 1: 0-7cm, layer 2: 7-28cm, and layer 3: 28-100cm). Additionally, two interface flux variables, mean surface latent heat flux (LHF) and mean surface sensible heat flux (SHF), are utilized to characterize thermodynamic coupling between Earth system components (*53*). A comprehensive list of these variables along with their abbreviations is provided in Supplementary Table S1. All variables are interpolated to $1.5° \times 1.5°$ spatial resolution ($121 \times 240$ global grid points) with daily temporal resolution derived from 1-hour ERA5 data through temporal averaging. The dataset spans from 1979 to 2021, partitioned into training (1979 to 2015), validation (2016 to 2017), and test (2018 to 2021) sets.

This study employs ECMWF S2S with model version CY48R1 reforecast data as the primary benchmark, comprising 11 ensemble members with twice-weekly initializations (Tuesdays and Fridays) providing forecasts up to 46 days. Additionally, we conduct comparative analysis with FuXi-S2S, a state-of-the-art deep learning-based ocean-atmosphere coupled prediction system that has demonstrated exceptional performance at subseasonal timescales up to 42 days. Both models are evaluated over the same period from 2018 to 2021. To quantify forecast anomalies, a 15-year climatology (2003 to 2017) is computed separately for each model using their respective reforecast data, with the seasonal cycle calculated using a 31-day centered moving window (±15 days) for each calendar day to obtain smooth and stable climatological references.

**TianXing-S2S overview**

TianXing-S2S is a probabilistic forecast model that provides global daily mean forecasts of 81 variables at 1.5° resolution. Following recent advances in autoregressive weather prediction (*2*, *4*), TianXing-S2S forecasts the next state from two consecutive historical states as

$$X^{t+1} = \text{TianXing-S2S}\left(X^t, X^{t-1}\right) \tag{1}$$

where $t$ represents the different sample time steps in the training set, $X^t$ denotes the complete set of all predictands at time $t$ (81 variables in total, detailed in the Datasets subsection), comprising 71 atmospheric variables (denoted as $A^t$) and 10 multi-sphere boundary condition variables (denoted as $B^t$). $A^t$ and $B^t$ are concatenated along the channel dimension as $X^t = \left[A^t, B^t\right]$ with shape (81, 121, 240) (121 latitude × 240 longitude grid points).

TianXing-S2S employs a two-stage Latent Diffusion Model (LDM) architecture including: Stage 1 compresses high-dimensional predictors into a compact latent space via VQVAE, while Stage 2 performs probabilistic prediction through diffusion processes, with an optimal transport (OT)-based coupling mechanism embedded within the denoising architecture to ensure physically plausible multi-sphere interactions. The following subsections detail these two stages.

**Stage 1: VQVAE Embedding**

The first stage of TianXing-S2S employs VQVAE, a discrete representation learning method to encode raw meteorological fields into a compact latent space. The motivation for this approach stems from the inherently discrete nature of many subseasonal phenomena, including positive/negative phases of ENSO events, monsoon active/break transitions, and blocking pattern formation/collapse. Through VQVAE encoding, the model extracts essential features while preserving modal characteristics crucial for subseasonal prediction.

Specifically, TianXing-S2S follows the standard VQVAE framework in (*33*), maintaining a codebook containing $K$ learnable entries, where each entry $e^k \in \mathbb{R}^d$ represents learned meteorological patterns or modes. In the embedding process, input fields are first encoded into feature space and then matched to their nearest codebook entries via Euclidean distance, transforming high-dimensional continuous meteorological data into compact discrete latent representations. Recognizing the distinct spatiotemporal scales and physical properties of atmospheric versus boundary variables, TianXing-S2S employs a grouped encoding strategy (Fig. 7 A), where atmospheric variables $A^t$ and boundary conditions $B^t$ at time step $t$ are processed through separate parameters to yield latent representations $ZA^t$ and $ZB^t$, respectively. The embedded shape in latent space has dimensions of

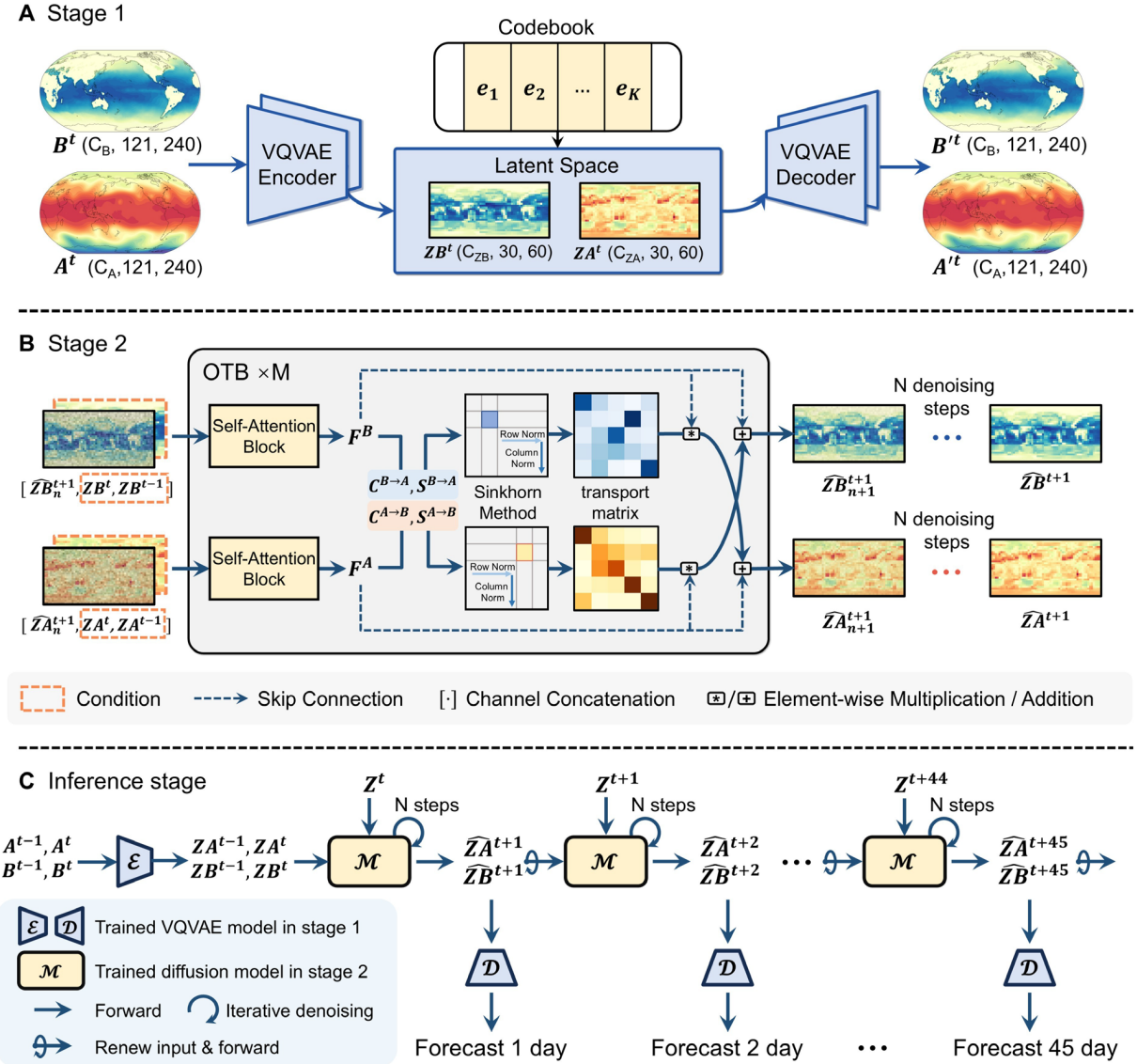

Fig. 6. The architecture and inference process of TianXing-S2S. (A and B) The two-stage architecture of TianXing-S2S. (A) Stage 1: VQVAE Embedding. $A^t$ and $B^t$ represent the atmospheric and boundary predictands of TianXing-S2S, which are encoded into latent space as $ZA^t$ and $ZB^t$ via $K$ learnable codebook entries, respectively. (B) Stage 2: Diffusion-based Probabilistic Modeling. $\widehat{ZA}_n^{t+1}$ and $\widehat{ZB}_n^{t+1}$ represent the noise-incorporated forecast for time step $t+1$ with diffusion step $n$ conditioned on two historical information. The optimal transport blocks (OTBs) formulate cross-sphere coupling as an OT problem and solve it using Sinkhorn method. The solved transport matrix theoretically represents the most efficient interaction pathway between atmospheric variables and multi-sphere boundary conditions. This denoising process executes N iterations to output the noise-free forecast $\widehat{ZA}^{t+1}$ and $\widehat{ZB}^{t+1}$. (C) Inference procedure after model training. Starting from two initial time steps, the model autoregressively generates forecasts by sampling from Gaussian noise, denoising through N steps, and renewing the input with predicted states. The process continues for 45 days with final decoding back to physical space at 1.5° resolution, producing a single ensemble member that can be repeated with different noise samples for ensemble forecasts.

30 × 60 grid points, with encoded feature dimensions $C_{ZA}$ and $C_{ZB}$ substantially compressed from their original dimensions $C_A$ (71 in this study) and $C_B$ (10 in this study), achieving significant dimensionality reduction while preserving essential information.

To ensure the model learns meaningful codebook entries and achieves high-quality reconstruction, TianXing-S2S follows the original VQVAE optimization strategy with the two-term loss function as

$$\mathcal{L}_{s1} = \mathrm{E}_t \left[ \left\| \mathcal{L}_{\text{rec}} + \mathcal{L}_{\text{codebook}} \right\|^2 \right] \tag{2}$$

where E represents the mathematical expectation. The reconstruction loss $\mathcal{L}_{\text{rec}}$ uses mean squared error (MSE) to measure discrepancies between original and reconstructed fields, ensuring the encoding-decoding process preserves essential physical information. The codebook loss $\mathcal{L}_{\text{codebook}}$ optimizes the alignment between embedded latent representations and codebook entries, enabling codebook vectors to accurately represent encoder output features. Taking atmospheric variable training as an example, the optimization target loss is

$$\mathcal{L}_{\text{rec}} = \mathrm{E}_{t,h,w} \left[ \omega_h \left\| A_{h,w}^t - A_{h,w}'^t \right\|^2 \right] \tag{3}$$

$$\mathcal{L}_{\text{codebook}} = \mathrm{E}_t \left[ \left\| \mathcal{E}(A^t) - z_q(A^t) \right\|^2 \right] \tag{4}$$

where $A_{h,w}'^t$ represents the reconstructed field after processing through VQVAE encoder $\mathcal{E}$ and decoder $\mathcal{D}$. The subscripts $h$ and $w$ denote the $h$-th longitudinal and $w$-th latitudinal grid point, $\omega_h$ denotes the latitude weighting following previous studies, $z_q(A^t)$ is the nearest codebook vector to the encoded output $\mathcal{E}(A^t)$. The same optimization procedure applies to boundary condition variables, with corresponding reconstruction and codebook losses computed using their respective encoders and decoders.

### Stage 2: Diffusion-based Probabilistic Modeling

Diffusion models have shown powerful generative capabilities with demonstrated success in data assimilation (*54*), probabilistic weather forecasting (*4, 55*), and super-resolution (*56, 57*). Unlike deterministic models that produce single forecasts, diffusion models learn the underlying data distribution to generate multiple plausible samples, enabling effective uncertainty representation in subseasonal prediction. Leveraging the latent representations from stage 1, TianXing-S2S then applies diffusion process in this compact space to generate ensemble forecasts.

The diffusion process learns the conditional probability distribution $p(ZX^{t+1} | ZX^t, ZX^{t-1})$, where $ZX^t = [ZA^t, ZB^t]$ represents the encoded latent representation at time $t$. Following the conventional denoise diffusion probabilistic model (DDPM) framework (*31*), the forward diffusion process gradually adds Gaussian noise as

$$ZX_n^t = \sqrt{\bar{\alpha}_n} \cdot ZX_0^t + \sqrt{1 - \bar{\alpha}_n} \cdot Z^t \tag{5}$$

where $n \in \{1, 2, \ldots, N\}$ denotes the diffusion steps, $ZX_n^t$ represents a noise-perturbed version of $ZX^t$ with progressively increasing noise levels of $n$, $ZX_N^t$ approaches pure Gaussian noise, and $ZX_0^t \coloneqq ZX^t$. $\bar{\alpha}_n$ is the predefined noise strength hyperparameter, $Z^t$ is standard Gaussian noise always sampled independently at time step $t$. As a counterpart, the reverse process learns to recover the target forecast iteratively starting from pure Gaussian noise $ZX_N^{t+1}$. Each denoising step can be expressed as

$$\widehat{ZX}_{n-1}^{t+1} = \mathcal{M}\left(\widehat{ZX}_{n}^{t+1}, n, ZX^{t}, ZX^{t-1}\right) \tag{6}$$

where $\mathcal{M}$ denotes the denoising network that predicts the latent state $\widehat{ZX}_{n-1}^{t+1}$ by progressively removing noise from $\widehat{ZX}_{n}^{t+1}$, conditioned on the two historical states and diffusion step. TianXing-S2S employs a UniPCMultiStep scheduler (*58*) to manage sampling process, achieving high-quality generation within 15 denoising iterations.

The design of the denoising network is crucial for generation quality. We propose OTB that facilitates more reasonable interaction between atmospheric and boundary condition variables (Fig. 7 B). Our approach recognizes that cross-sphere coupling interactions, though governed by complex and unresolved physical mechanisms, manifest as statistically dominant modes that constitute the primary source of subseasonal predictive skill. Dominant modes refer to statistically robust patterns through which cross-sphere interactions systematically influence atmospheric forecasts at subseasonal timescales. For instance, tropical sea surface temperature anomalies influence mid-to-high latitudes through well-established teleconnection patterns such as the Pacific-North American (PNA) pattern (*59*) and North Atlantic Oscillation (NAO) (*60*). Similarly, Madden-Julian Oscillation (MJO) propagation over the Maritime Continent depends critically on air-sea coupling strength (*61*), while monsoon moisture transport operates predominantly through specific atmospheric rivers and low-level jets (*62*). These examples demonstrate that cross-spherical signal transmission at subseasonal timescales follows statistically organized pathways rather than random interactions. We therefore formulate cross-spherical coupling as probability distributions and incorporate OT theory (*29*) to model these dominant modes. Based on these dominant modes, we then employ the Sinkhorn algorithm to solve for an interaction matrix (*63*, *64*). Theoretically, this interaction matrix ensures the most efficient and effective information transfer between atmospheric variables and multi-sphere boundary conditions.

Specifically, the denoiser separates noisy input $\widehat{ZX}_{n}^{t+1}$ and conditional inputs $\left[ZX^{t}, ZX^{t-1}\right]$ along the channel dimension into atmospheric $\left[\widehat{ZA}_{n}^{t+1}, ZA^{t}, ZA^{t-1}\right]$ and boundary $\left[\widehat{ZB}_{n}^{t+1}, ZB^{t}, ZB^{t-1}\right]$ components and passes them through a sequence of OTBs. Each block utilizes independent self-attention modules to extract atmospheric features $F^{A}$ and boundary condition features $F^{B}$, capturing distinct spatiotemporal patterns across different Earth system components. Based on the framework of OT, we then calculate a cost matrix C by grid-point-wise cosine similarity between atmospheric features $F^{A}$ and boundary condition features $F^{B}$, quantifying the preference for specific coupling pathways. The optimal transport matrix (OTM) S is initialized according to C and iteratively solved using the Sinkhorn algorithm (*63*, *64*), with each element representing the coupling strength between corresponding atmospheric and boundary grid points. Different sets of parameters are employed to learn $C^{B \to A}, S^{B \to A}$ and $C^{A \to B}, S^{A \to B}$ independently, as atmosphere-to-boundary and boundary-to-atmosphere influences follow distinct physical mechanisms. Cross-spherical coupling is ultimately achieved through element-wise multiplication between the transport matrices and corresponding features, followed by residual connections to preserve the original feature information.

During training, we optimize the denoising network to fit the forward noising process output as

$$\mathcal{L}_{s2} = \mathrm{E}_{t,n}\left[\left\|ZX_{n-1}^{t+1} - \mathcal{M}\left(\widehat{ZX}_{n}^{t+1}, n, ZX^{t}, ZX^{t-1}\right)\right\|^{2}\right] \tag{7}$$

where $n$ follows a uniform distribution over {1, 2, ..., N}. Latitude-weighted loss is not applied since optimization is performed in the latent space.

**Training details**

TianXing-S2S is implemented with the PyTorch framework and trained on 8 NVIDIA A800 GPUs (80GB memory). The training procedure consists of two stages corresponding to the hierarchical model architecture. In stage 1, the VQVAEs are trained using the AdamW optimizer (*65*) with $\beta_1$ = 0.9, $\beta_2$ = 0.99, and a batch size of 4 for 200 epochs. The learning rate is initialized at $1\times10^{-4}$ and decayed to $5\times10^{-6}$ via cosine annealing, with training converging taking about 10 days. The diffusion model stage 2 is first trained for single-step prediction with 150 epochs using AdamW with $\beta_1$ = 0.9, $\beta_2$ = 0.95, a batch size of 8, and learning rate decaying from $1\times10^{-4}$ to $1\times10^{-5}$. Subsequently, we adopt an autoregressive training strategy with replay buffer as in (*66*) for multi-step fine-tuning, progressively increasing the forecast horizon from 1 to 10 steps. Each step undergoes 30 epochs of training with a constant learning rate of $2\times10^{-5}$ and batch size of 1 due to memory constraints. This autoregressive strategy effectively mitigates long-time error accumulation. The complete training process of stage 2 requires approximately 14 days in total.

**Inference stage**

The fully trained TianXing-S2S generates forecasts through an autoregressive inference procedure (Fig. 7 C). Initially, two consecutive time steps of atmospheric and boundary condition variables are encoded into the latent space via the trained VQVAEs to serve as initial conditions. At each forecast step, a sample is drawn from the standard Gaussian distribution and refined through N denoising iterations to produce the next time step prediction in the latent space. The predicted latent state then renews the older time step in the input conditions, forming a sliding window that continues autoregressively forecast. This process iteratively generates predictions up to 45 days (up to 180 in long-term simulation experiments). Finally, the latent space predictions are decoded back to physical space using the VQVAE decoders, yielding full-field atmospheric and boundary condition forecasts at the original 1.5° × 1.5° resolution. The entire sampling and inference procedure is repeated multiple times with different random noise samples, generating an ensemble of forecast members.

**Evaluation metrics**

The performance of our model is statistically assessed using probabilistic metrics (CRPS and SSR for ensemble distribution reliability), deterministic metrics (RMSE and ACC for ensemble mean accuracy), and extreme metrics (BSS for tail-end coverage). Attribution analysis of extreme event prediction utilizes PIM and GradCAM methods. We also introduce the weighted mean influencing distance (WMID), a metric designed to quantify the spatial extent of cross-sphere coupling processes captured by the OTB. Additionally, MJO forecast skill is evaluated in the Supplementary Materials using the Real-time Multivariate MJO (RMM) index and the OLR-based MJO Index (OMI), which capture dynamical-convective structure and core convective signals, respectively.

## Acknowledgments

We acknowledge the ECMWF for providing access to the ERA5 reanalysis dataset and the S2S reforecast database, which are fundamental to this research. We thank the National Oceanic and Atmospheric Administration Physical Sciences Laboratory (NOAA PSL) for making available the OMI EOF patterns data that enabled our analyses. We are grateful to the Fuxi-S2S research team for making their forecast data publicly accessible. The availability and quality of these datasets are instrumental in facilitating the comprehensive analyses presented in this study.

**Funding:** This study is supported in part by the Meteorological Joint Funds of the National Natural Science Foundation of China under Grant U2542211 and U2542212, in part by the Original Exploration Project of the National Natural Science Foundation of China under Grant 42450163, and in part by the National Natural Science Foundation of China under Grant 42405147.

**Author contributions:** Conceptualization: Y.C. and B.M. Methodology: Y.C., H.G., and B.Q. Investigation: Y.C., B.M., H.G., and B.Q. Supervision: S.Y. and B.M. Writing—original draft: Y.C., B.M., S.Y, B.Q., and H.G. Writing—review & editing: Y.C., B.M., S.Y, B.Q., and H.G.

**Competing interests:** Authors declare that they have no competing interests.

**Data and materials availability:** The ERA5 dataset is available from the Copernicus Climate Data Store at https://cds.climate.copernicus.eu/. The ECMWF S2S dataset is accessed from https://apps.ecmwf.int/datasets/data/s2s/. The observed OMI EOF patterns data is obtained from NOAA PSL at https://psl.noaa.gov/mjo/mjoindex/. The FuXi-S2S data is downloaded from https://huggingface.co/datasets/FudanFuXi/FuXi-S2S. The source data of TianXing-S2S can be found in https://huggingface.co/yuxuan9909/datasets.

# Supplementary Materials for

## Skillful Subseasonal-to-Seasonal Forecasting of Extreme Events with a Multi-Sphere Coupled Probabilistic Model

Bin Mu *et al.*

*Corresponding author. Email: yuanshijin@tongji.edu.cn (S.Y.); brunoqin@163.com (B.Q.)

**This PDF file includes:**

Figs. S1 to S13
Table S1

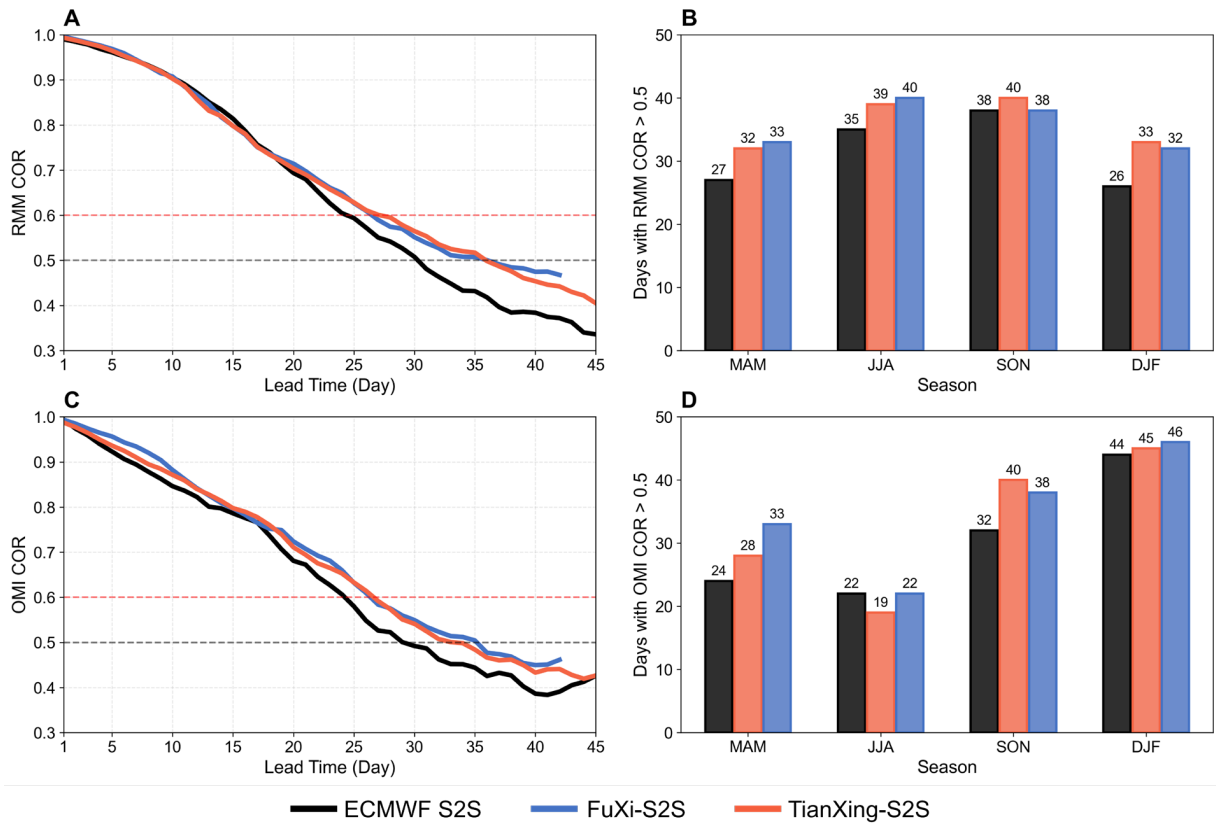

**Fig. S1. Forecast skill comparison of MJO bivariate correlation coefficient (COR). (A)** Comparisons of real-time multivariate COR with different forecast lead time for ECMWF S2S, FuXi-S2S, and TianXing-S2S over 2018 to 2021, with red and gray dashed lines indicating thresholds of 0.6 and 0.5, respectively. **(B)** Seasonal distribution of forecast lead times (days) maintaining RMM COR above 0.5, where seasonal labels represent three-month periods (e.g., MAM represents March-April-May for boreal spring). **(C)** Same as (A), but for COR defined by the OLR-based MJO Index (OMI). **(D)** Same as (B), but for OMI-based forecast skill duration.

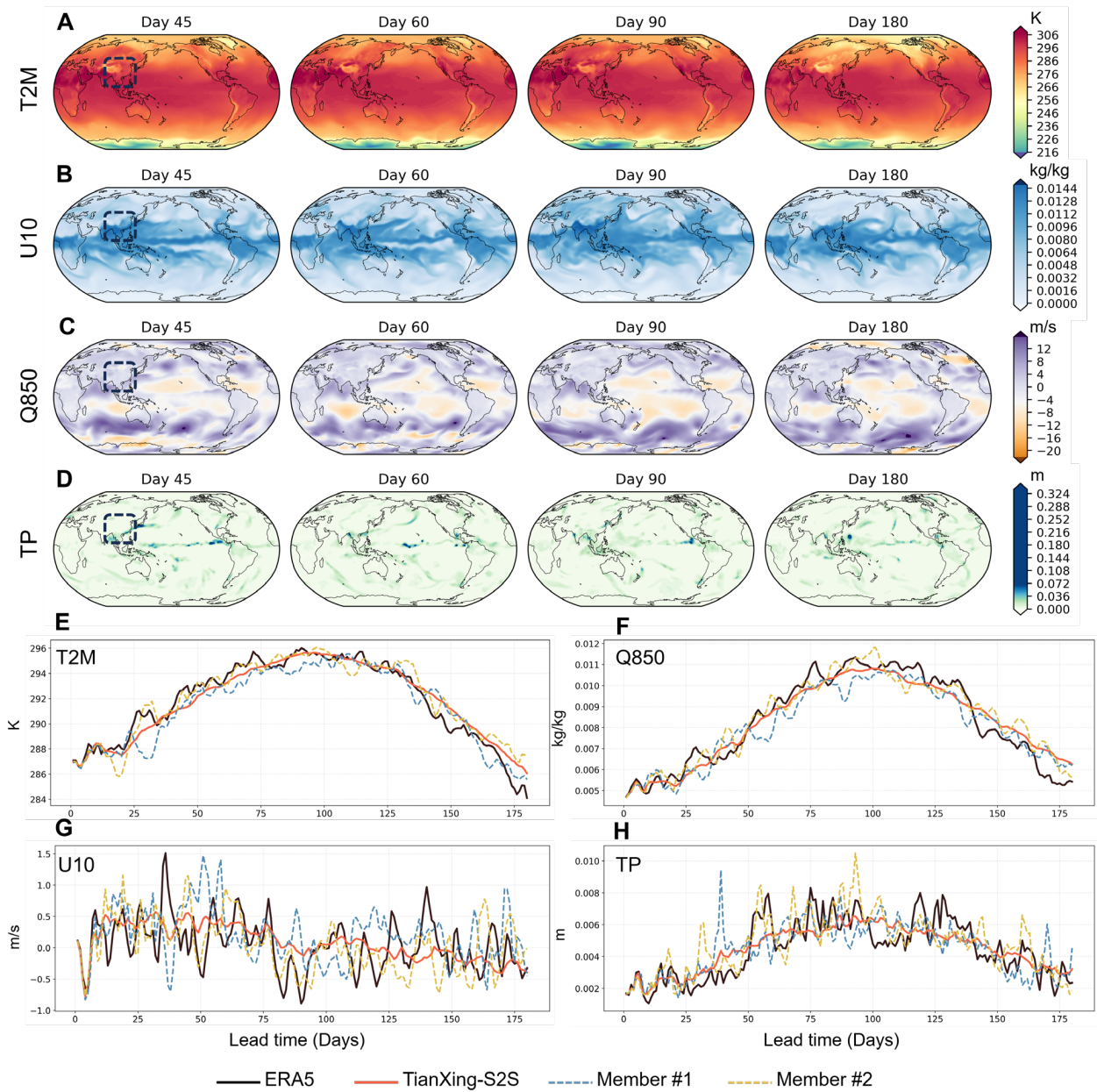

Fig. S2. Long-term forecasts up to 180 days from TianXing-S2S initialized on April 18, 2018, showing T2M, Q850, U10, and TP variables. (A to D) Global spatial patterns from TianXing-S2S ensemble member #1 (random selected) at forecast days 45, 60, 90, and 180 for T2M, Q850, U10, and TP, respectively. (E to H) Region-averaged time series for the dashed-box regions highlighted in the first column in panels (A to D), showing 1 to 180 days forecasts for T2M, Q850, U10, and TP. Ensemble mean is depicted as red solid lines, two randomly selected ensemble members as dashed lines, and ERA5 reanalysis as black lines for comparison.

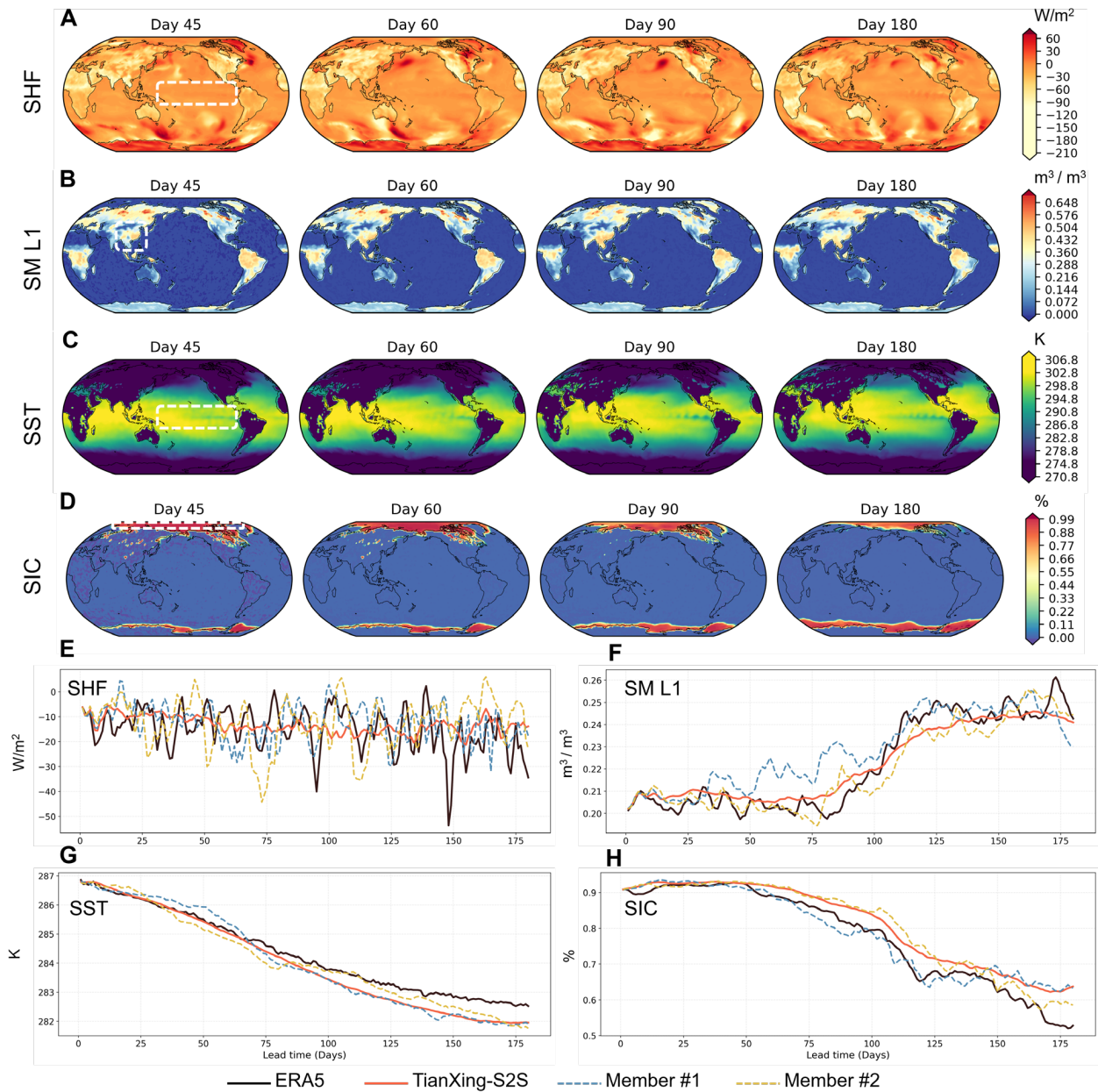

**Fig. S3. Same to Fig. S2., but for four boundary condition variables SHF, SM L1, SST, and SIC initialized on March 14, 2018.**

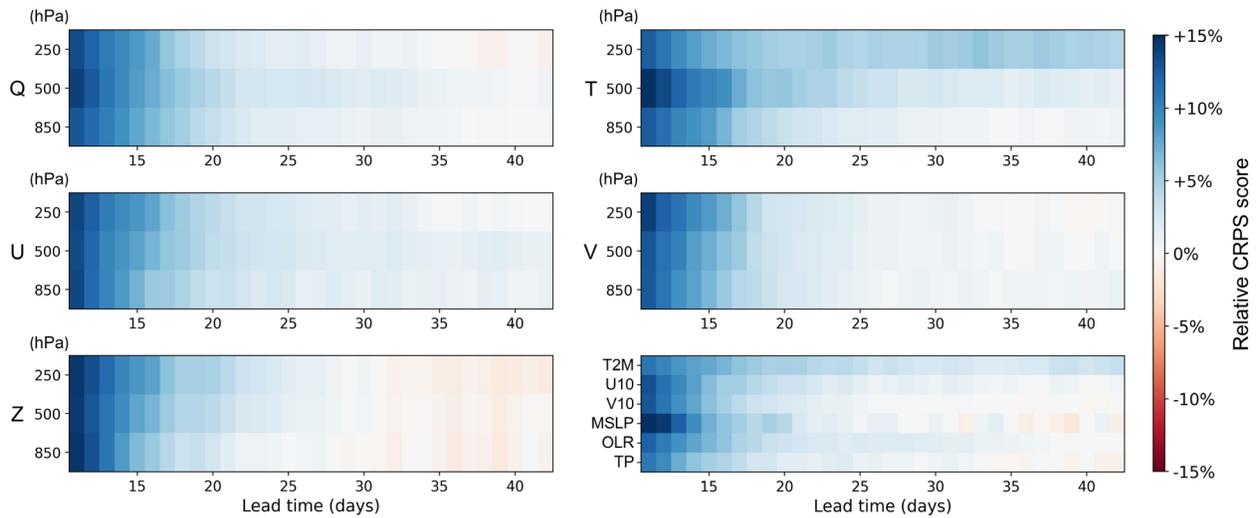

**Fig. S4. CRPS scorecard for TianXing-S2S versus FuXi-S2S with 42-day lead time averaged over 2018-2021.** Each subplot displays CRPS scores for a variable category across different forecast lead times, with each row representing atmospheric pressure level variables or surface variables. Blue shading indicates TianXing-S2S outperforms FuXi-S2S, red shading indicates the opposite, and values approaching white represent comparable performance between the two models.

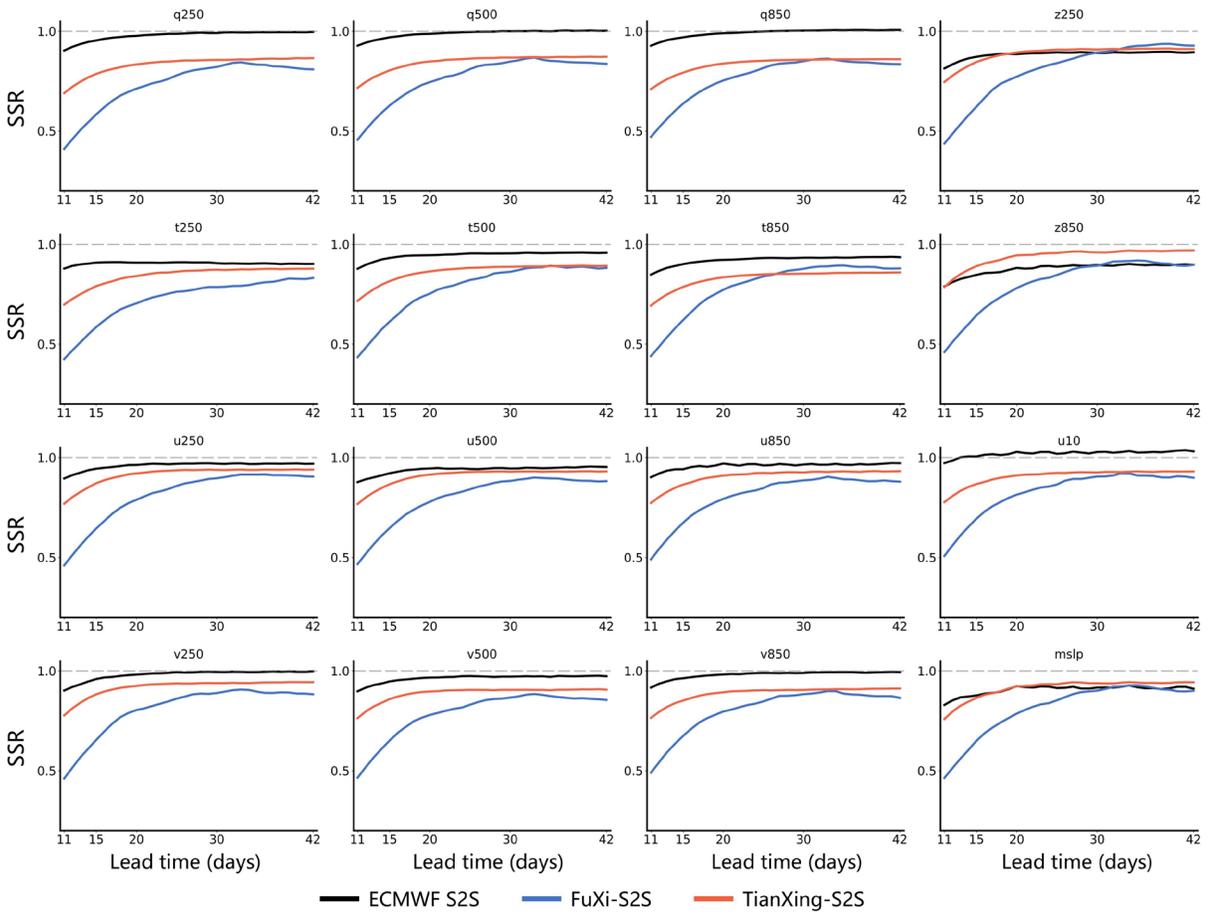

**Fig. S5. SSR for additional selected variables with different forecast lead times averaged over 2018-2021.** The dashed line represents the threshold of 1, which is the theoretical optimum.

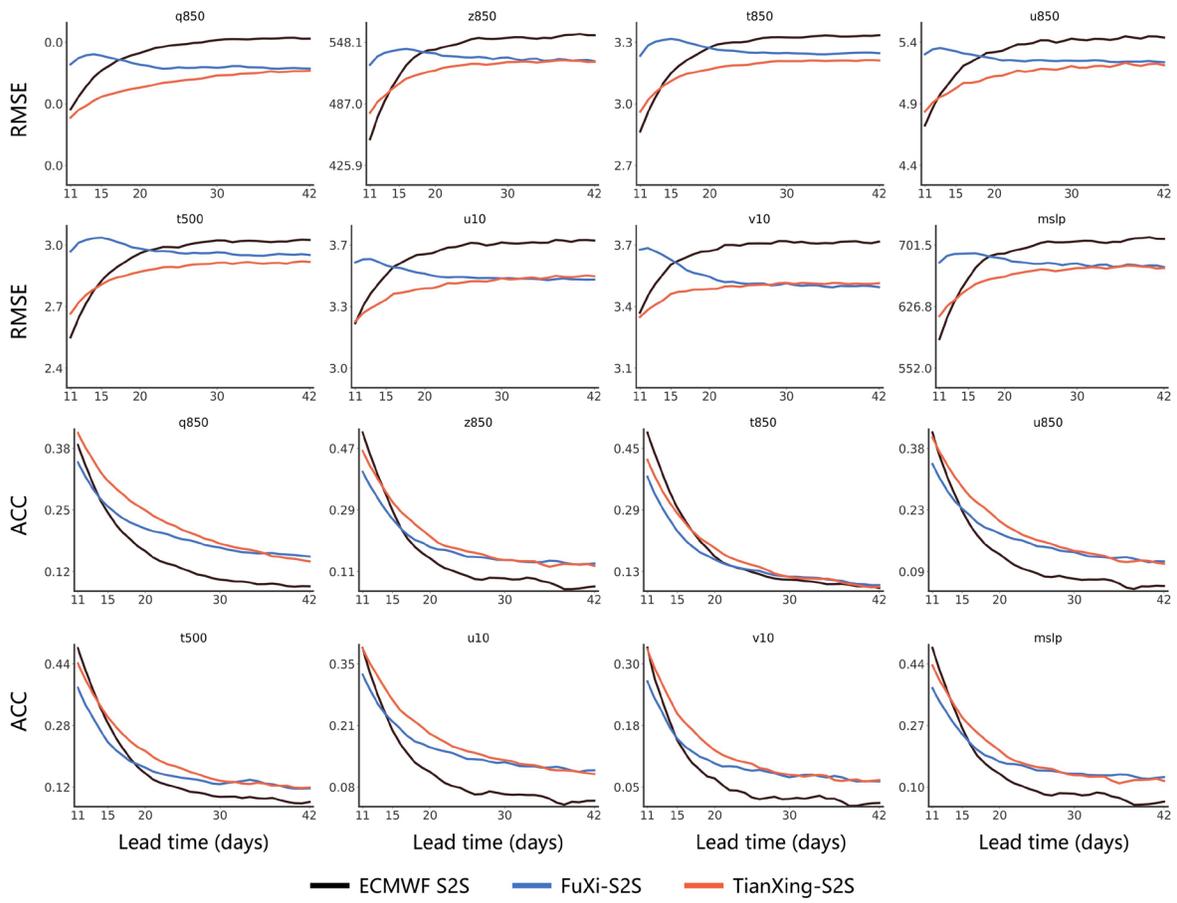

**Fig. S6. RMSE and ACC for additional selected variables with different forecast lead times averaged over 2018-2021.**

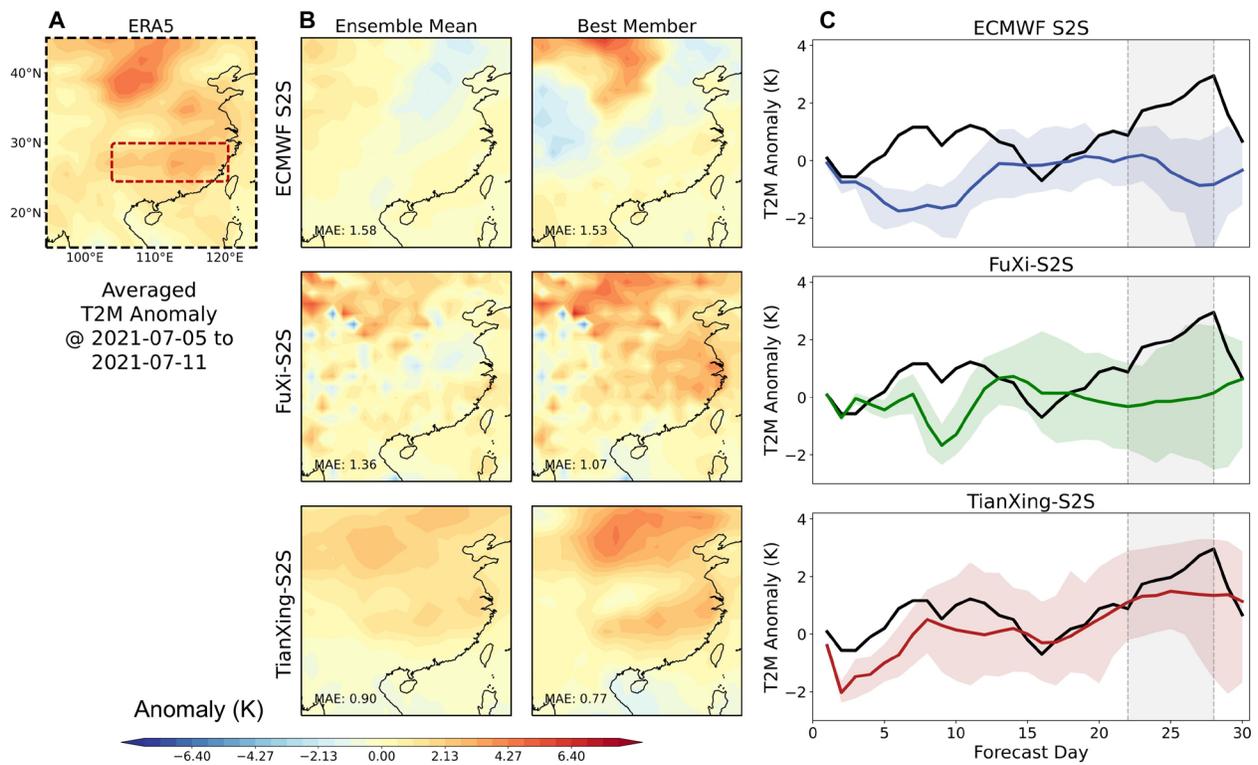

**Fig. S7. Comparison of East Asian extreme heat wave prediction during July 5 to 11, 2021 by TianXing-S2S, FuXi-S2S, and ECMWF S2S.** (A) Mean T2M anomalies over East Asia during July 5 to 11, 2021, showing extreme positive temperature anomalies over Southeast region of China, highlighted by the red dashed rounded rectangle. (B) Ensemble mean (the first column) and best forecast member (the second column) of the heat wave event from ECMWF S2S (the first row), FuXi-S2S (the second row), and TianXing-S2S (the third row) initialized 4 weeks earlier. The text displayed in the lower left corner of each subplot indicates the average absolute error over the whole field of the forecast. (C) Comparison of area-averaged temperature anomalies over the region outlined by the red rectangle in (A) with different forecast lead time. From top to bottom, the panels show the forecast results of ECMWF S2S, FuXi-S2S, and TianXing-S2S, respectively. The black solid lines represent ERA5 reanalysis values, the colored lines represent model forecast values with the colored shading indicates the coverage of ensemble members. The gray shading highlights the period when the extreme event occurred during the fourth week.

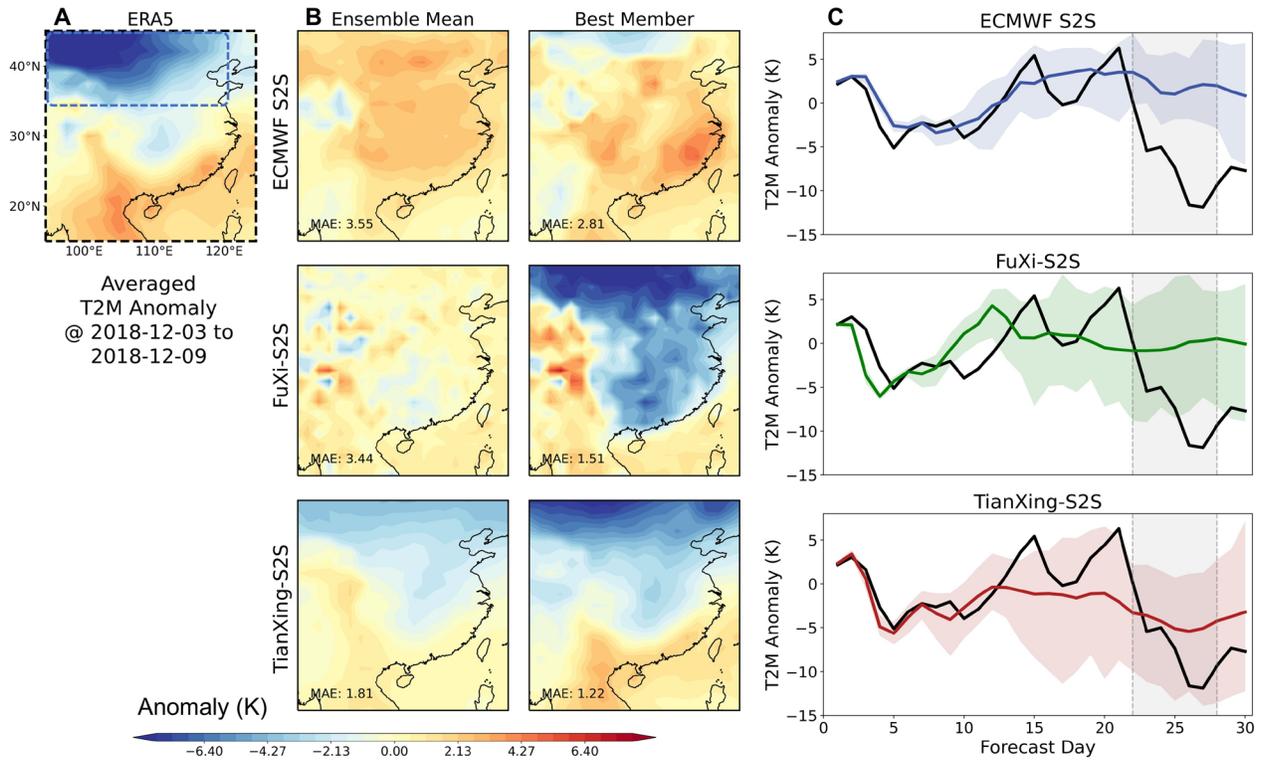

**Fig. S8.** Same to Fig. S7., but for the cold spell event during December 3 to 9, 2018.

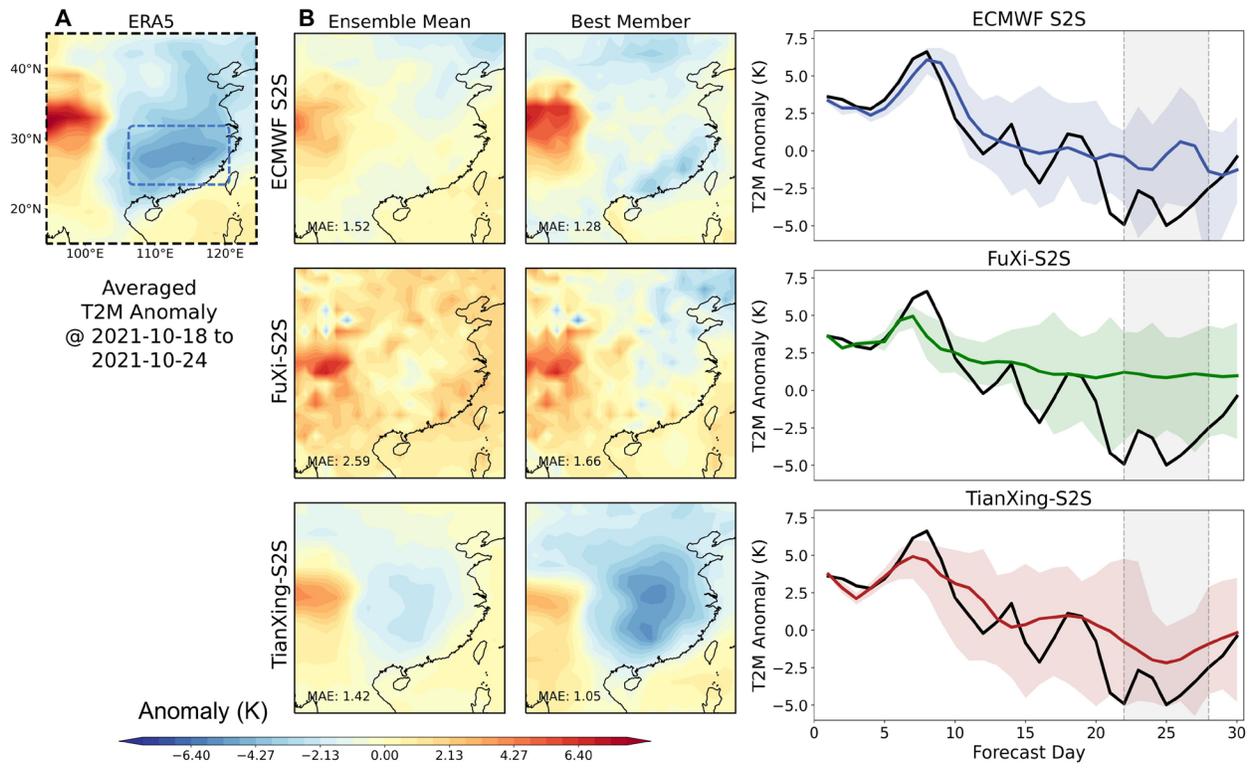

**Fig. S9.** Same to Fig. S7., but for the cold spell event during October 18 to 24, 2021.

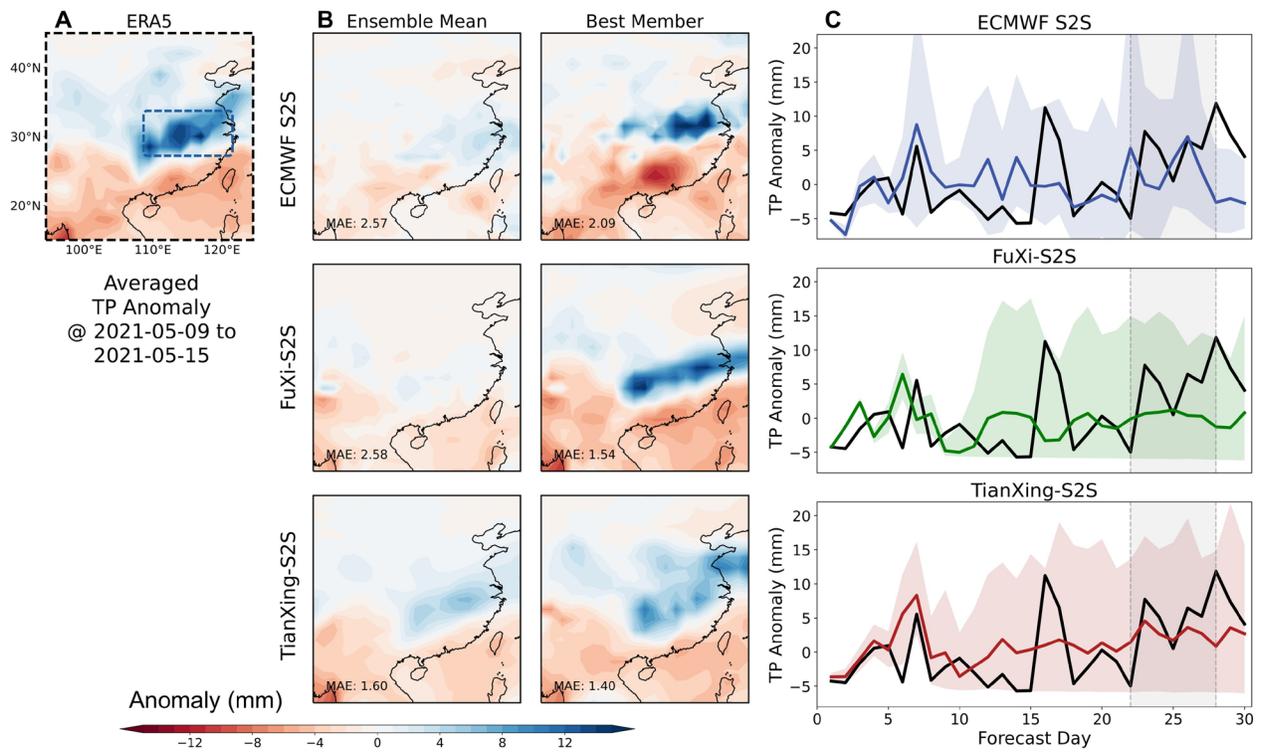

**Fig. S10.** Same to Fig. S7., but for the precipitation event during May 9 to 15, 2021.

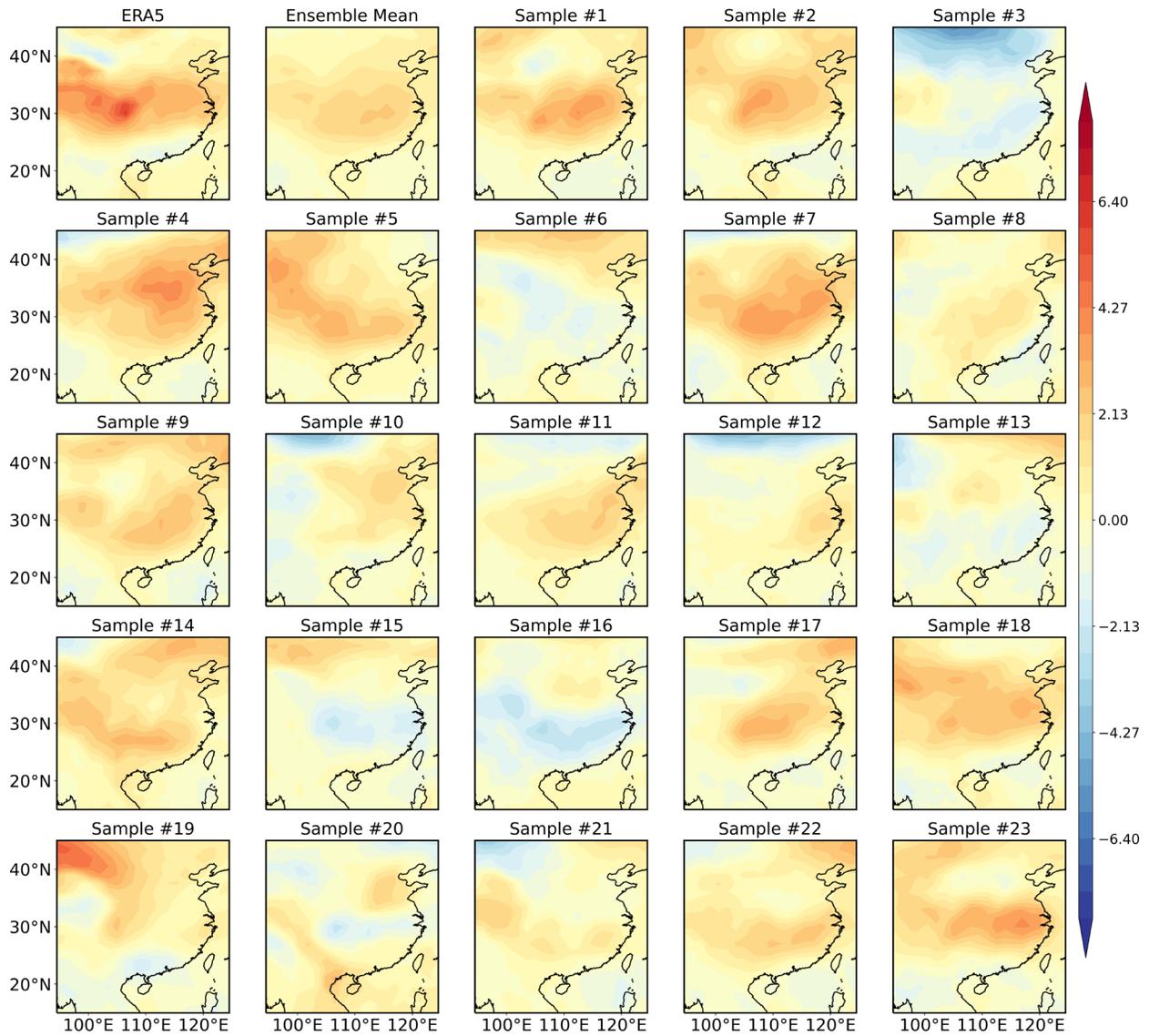

**Fig. S11. Visualization of selected ensemble members from TianXing-S2S for T2M anomaly prediction of the East Asian extreme heat wave event during August 29 to September 4, 2018.** The first row, first column shows ERA5 values, the first row, second column displays the ensemble mean, and the remaining subplots present 23 randomly selected ensemble members from the total 51 ensemble members.

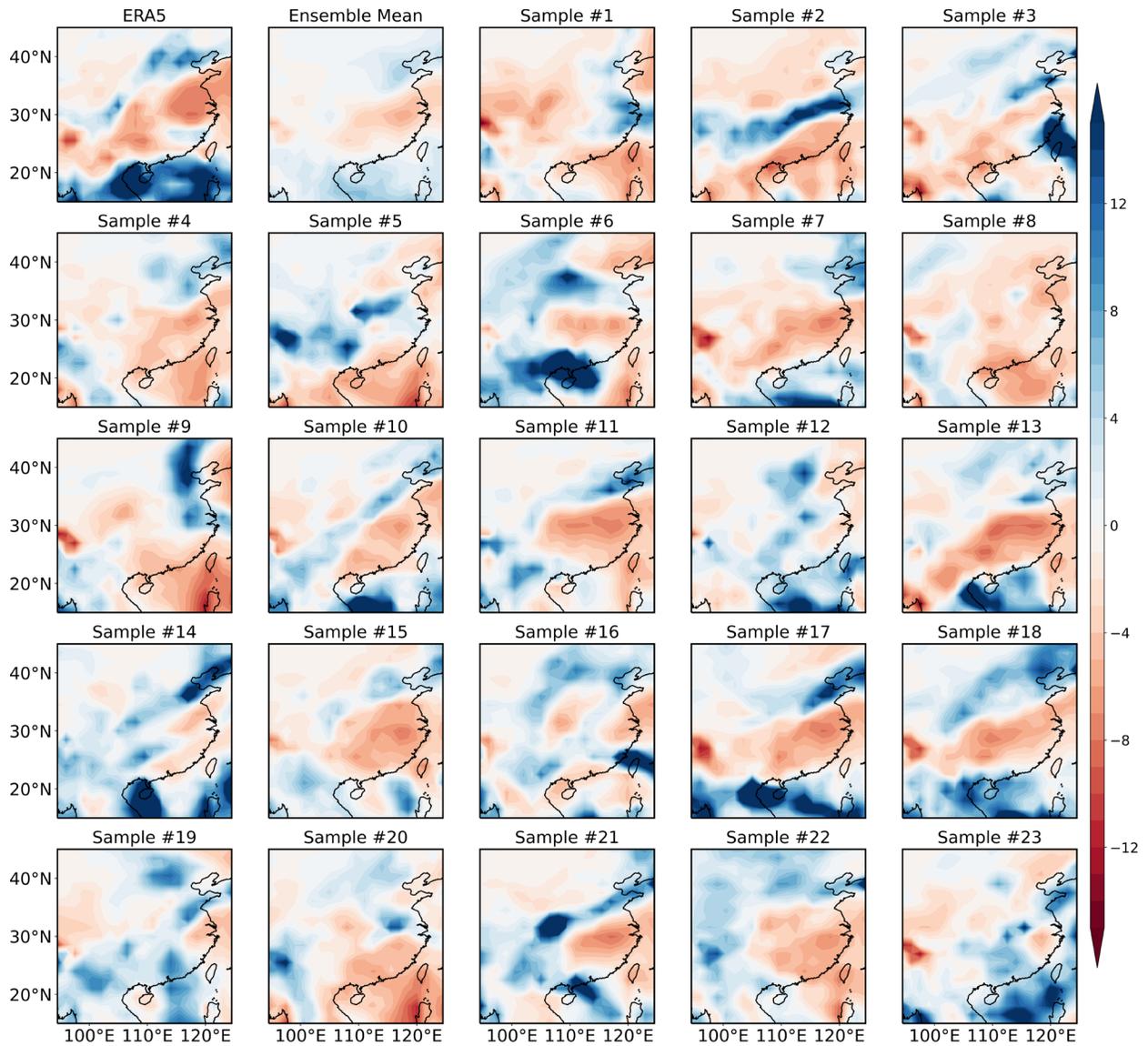

**Fig. S12. Same to Fig. S4., but for precipitation case during July 11 to 17, 2018.**

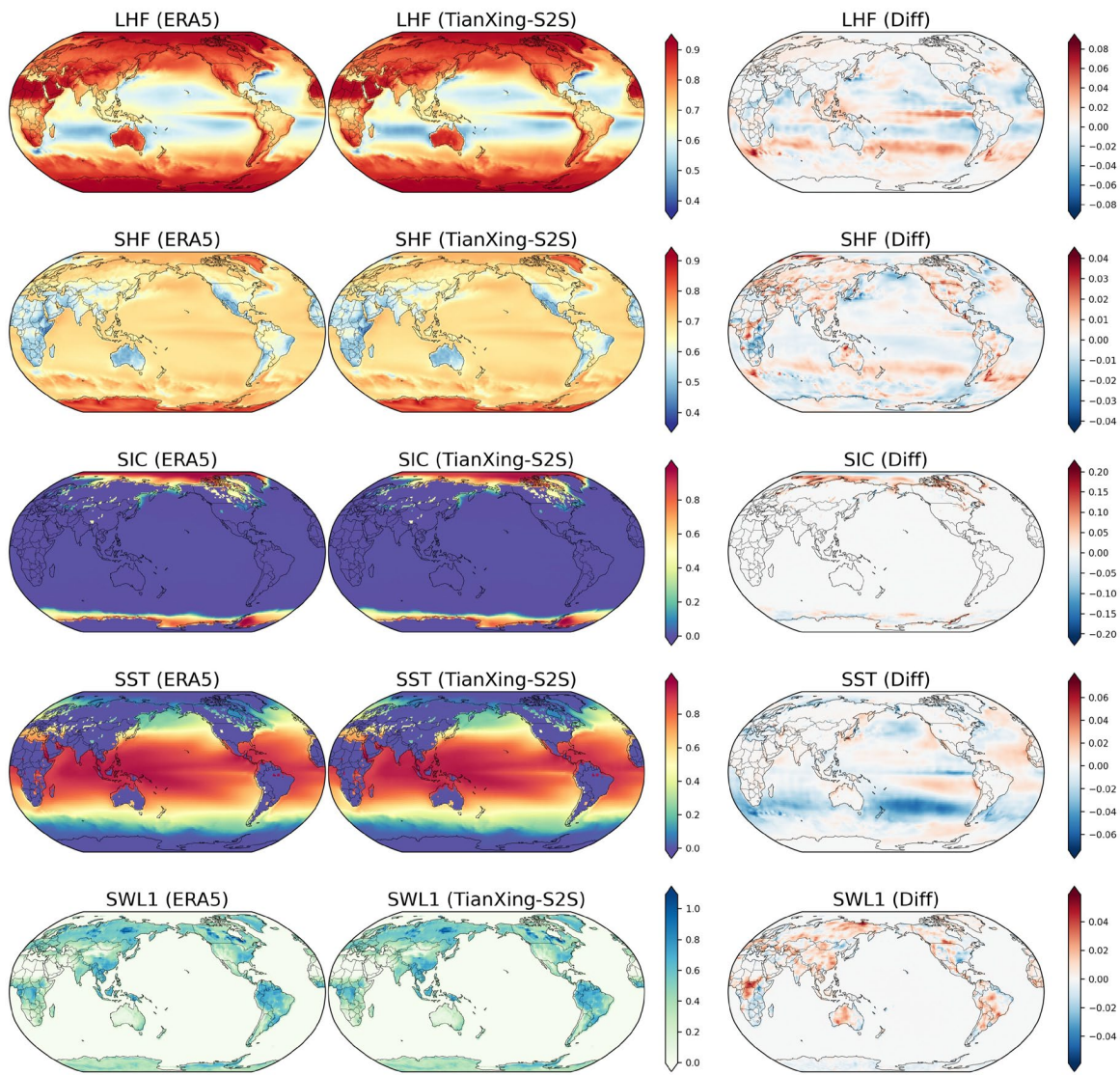

**Fig. S13. Climatological differences between TianXing-S2S multi-sphere boundary condition forecasts and ERA5 reanalysis.** The climatology is computed from daily averages over weeks 3-6 across the 2018-2021 evaluation period. Values on the colorbar represent min-max normalized results scaled to the range [0, 1].

**Table S1. Multi-sphere variables used in TianXing-S2S with their corresponding abbreviations.** All variables serve as both predictors and predictands in this study.

| Sphere | Level | Variable | Abbreviation |
|---|---|---|---|
| Atmosphere | Pressure levels at 50, 100, 150, 200, 250, 300, 400, 500, 600, 700, 850, 925, and 1000 hPa | specific humidity | Q |
| | | temperature | T |
| | | u component of wind | U |
| | | v component of wind | V |
| | | geopotential | Z |
| | Single levels | 2-meter temperature | T2M |
| | | outgoing longwave radiation | OLR |
| | | total precipitation | TP |
| | | mean sea level pressure | MSLP |
| | | 10-meter u wind component | U10 |
| | | 10-meter v wind component | V10 |
| Ocean | Single levels | sea surface temperature | SST |
| | | sea-ice cover | SIC |
| Land | Depth levels at 0-7, 7-28, and 28-100 cm | soil temperature | ST |
| | | soil moisture | SM |
| Interface flux | Single levels | mean surface latent heat flux | LHF |
| | | mean surface sensible heat flux | SHF |